\documentclass[sn-mathphys-num]{sn-jnl}

\usepackage[utf8]{inputenc}
\usepackage{graphicx}%
\usepackage{multirow}%
\usepackage{amsmath,amssymb,amsfonts}%
\usepackage{amsthm}%
\usepackage{mathrsfs}%
\usepackage[title]{appendix}%
\usepackage{xcolor}%
\usepackage{textcomp}%
\usepackage{manyfoot}%
\usepackage{booktabs}%
\usepackage{algorithm}%
\usepackage{algorithmicx}%
\usepackage{algpseudocode}%
\usepackage{listings}%

\usepackage{lineno,hyperref}
\usepackage{epstopdf}
\usepackage{epsfig}
\usepackage{bm}
\usepackage{url}
\usepackage{threeparttable}
\usepackage{subfigure,color}
\usepackage{ragged2e}
\usepackage{soul}
\usepackage[figuresright]{rotating}
\usepackage{pdflscape}
\usepackage{moresize}
\usepackage{makecell}
\usepackage{tabularx}

\theoremstyle{thmstyleone}%
\newtheorem{theorem}{Theorem}
\theoremstyle{thmstyletwo}%

\theoremstyle{thmstylethree}%
\newtheorem{assumption}{Assumption}

\newcommand{\beq}{\begin{equation}}
\newcommand{\eeq}{\end{equation}}
\newcommand{\beqa}{\begin{eqnarray}}
\newcommand{\eeqa}{\end{eqnarray}}
\newcommand{\beqas}{\begin{eqnarray*}}
\newcommand{\eeqas}{\end{eqnarray*}}
\newcommand{\ba}{\begin{array}}
\newcommand{\ea}{\end{array}}
\newcommand{\bi}{\begin{itemize}}
\newcommand{\ei}{\end{itemize}}

\def\Arg{{\rm Arg}}
\def\red#1{\textcolor{red}{#1}}

\begin{document}

\title[Joint Learning for MC]{A Joint Sparse Self-Representation Learning Method for Multiview Clustering}



\author[1]{\fnm{Mengxue} \sur{Jia}}\email{jmgxd@stu.xidian.edu.cn}

\author*[1]{\fnm{Zhihua} \sur{Allen-Zhao}}\email{allenzhaozh@gmail.com}

\author[2]{\fnm{You} \sur{Zhao}}\email{Zhaoyou1991sdtz@163.com}

\author[1]{\fnm{Sanyang} \sur{Liu}}\email{liusanyang@126.com}


\affil[1]{\orgdiv{School of Mathematics and Statistics}, \orgname{Xidian University}, \orgaddress{\city{Xi'an}, \postcode{710126}, \state{Shaanxi}, \country{China}}}

\affil[2]{\orgdiv{Chongqing Key Laboratory of Nonlinear Circuits and Intelligent Information Processing, College of Electronic and Information Engineering},
\orgname{Southwest University}, \orgaddress{\city{Chongqing}, \postcode{400715}, \country{China}}}



\abstract{
Multiview clustering (MC) aims to group samples using consistent and complementary information across various views.
The subspace clustering, as a fundamental technique of MC, has attracted significant attention.
In this paper, we propose a novel joint sparse self-representation learning model for MC,
where a featured difference is the extraction of view-specific local information
by introducing cardinality (i.e., $\ell_0$-norm) constraints instead of Graph-Laplacian regularization.
Specifically, under each view, cardinality constraints directly restrict the samples used in the self-representation stage to extract reliable local and global structure information,
while the low-rank constraint aids in revealing a global coherent structure in the consensus affinity matrix during merging.
The attendant challenge is that Augmented Lagrange Method (ALM)-based alternating minimization algorithms cannot guarantee convergence
when applied directly to our nonconvex, nonsmooth model, thus resulting in poor generalization ability.
To address it, we develop an alternating quadratic penalty (AQP) method with global convergence,
where two subproblems are iteratively solved by closed-form solutions.
Empirical results on six standard datasets demonstrate the superiority of our model
and AQP method, compared to eight state-of-the-art algorithms.}




\maketitle

\section{Introduction}\label{sec1}

In the information age, the prevalence of multiview data is evident.
For instance, an object or concept can be described through various mediums
such as words, pictures, and videos, while a news event can be reported in
multiple languages.
How to extract valid information from the multiview data is a worthy problem.
Multiview clustering (MC) has been a crucial technique for handling multiview data,
which is successfully applied in various fields
such as machine learning and pattern recognition \cite{bmvc},
object recognition \cite{appliedin2}, text clustering \cite{clusteringText}.

The subspace clustering 
\cite{multiview-subspace}, as a fundamental technique of MC,
has attracted significant attention.
The relevant methods can be categorized as the latent- and self-representation ones \cite{glmsc,jsmc},
where the former aims to discover shared latent representations
across all views, based on which the self-representation is performed,
while the latter directly performs the self-representation on the original data
for capturing the structure information under each view.
By means of these structure information,
a final affinity matrix is generated for clustering.
Therefore, the multiview subspace methods generally hold two stages:
the view-specific information learning stage and the consensus information learning stage.
In this view, we roughly divide the self-representation methods of multiview subspace clustering
into two types of approaches.
\textit{The first one} attempts to find sufficiently good view-specific affinity matrices $C^v$,
whose generalized model is formulated as
\begin{equation}
    \label{generalfd}
    \begin{aligned}
        \min\quad &\sum_{v=1}^{V} \left\{ \mathcal{L}(X^v,X^vC^v) + \mathcal{R}_1(C^v) \right\} + \phi (C^1, \cdots, C^V) \\
        \text{s.t.}\quad &C^v\in\Omega(C^v),\ v \in [V],
    \end{aligned}
\end{equation}
and then obtain the consensus affinity matrix $C^*$
by the average or the sum of all the view-specific affinity matrices.
\textit{The second one} aims to simultaneously enhance both view-specific affinity matrices and consensus affinity matrix within a unified model, written as
\begin{equation}
    \label{generalsd}
    \begin{aligned}
        \min\quad &\sum_{v=1}^{V}\left\{ \mathcal{L}(X^v,X^vC^v) + \mathcal{R}_1 (C^v)\right\}  + \sum_{v=1}^{V} \varphi (C^v, C^*) +\mathcal{R}_2 (C^*) \\
        \text{s.t.}\quad &C^v\in\Omega(C^v),\  v \in [V]; \ C^*\in\Omega(C^*).
    \end{aligned}
\end{equation}
In \eqref{generalfd} and \eqref{generalsd},
$\mathcal{L}(\cdot,\cdot)$ represents the loss function;
$\mathcal{R}_1(\cdot)$ and $\mathcal{R}_2(\cdot)$ are the regularization terms;
$\phi(C^1, \cdots, C^V)$ is the cross-term among all the views;
$\varphi(C^v,C^*)$ is the learning term of consensus information;
$\Omega(\cdot)$ is the relevant feasible set.

During the self-representation process,
original samples reconstruct each other to comprehensively capture the global structure \cite{subspacekind}. However, relying solely on global structure information is inadequate for good clustering performance.
Therefore, integrating local information during self-representation is a critical issue.
Most existing methods incorporate local information via either the smoothness term or the rank of the Laplacian matrix.
The key distinction between these two methods is that the Laplacian matrix in the smoothness term is computed from the original data, retaining the original local structure,
while the Laplacian matrix in the rank constraint is derived from the learned affinity matrix to ensure the learned information has a local structure.
For instance, within the framework of model \eqref{generalfd},
Cao et al. \cite{cvpr2015} adopted the trace of the Laplacian ellipsoidal norm of each view-specific affinity matrix as the smoothness term,
and then used the Hilbert-Schmidt Independence Criterion (HSIC) to extract the diversity among views.
Based on model \eqref{generalsd},
Zheng et al. \cite{neuro2020} and Cai et al. \cite{jsmc} exploited the Laplacian matrices of all views to generate the smoothness terms on the consensus affinity matrix.
The difference between them lies in that the former implemented the self-representation model in a two-layer model, while the latter assumed that the view-specific affinity matrix was composed of two components: the view noise part and the consensus part.
Furthermore, by leveraging the property of the rank of the Laplacian matrix \cite{Kyfan},
Zhang et al. \cite{Zhang2022} derived a unified low-dimensional representation
by imposing a Laplacian-rank constraint on each view-specific affinity matrix.
In a similar vein, Qin et al. \cite{subspaceKinds} applied rank constraints to both view-specific and consensus affinity matrices, aiming to equate the ranks of Laplacian matrices associated with them.

Beyond local structure information, affinity matrices can significantly enhance clustering performance
when endowed with some properties.
Sparse and low-rank characteristics, in particular, are two properties widely exploited in this context.
In line with model \eqref{generalfd}, Yin et al. \cite{neuro2015} utilized the $\ell_1$ norm to generate sparse affinity matrices.
Additionally, they designed a cross-term of the $\ell_1$ view information to achieve group-sparse effects.
Chen et al. \cite{mlrr} employed the nuclear norm to pursue low-rank effects on view-specific affinity matrices,
where the cross-terms among all the views were formulated as the Frobenius norms on these view-specific affinity matrices.
Brbi{\'c} and Kopriva \cite{pr2018} simultaneously exploited the low-rank and sparse properties on the view-specific affinity matrices by using the nuclear norm and the $\ell_1$ norm.
Moreover, they adopted the Frobenius norm for the cross-terms as a part of the loss function.
Guo et al. \cite{tpami2023} applied the tensor logarithmic Schatten-$p$ norm to the tensor composed of all view-specific affinity matrices to obtain the low-rank property.
Under model \eqref{generalsd}, Cai et al. \cite{jsmc} minimized the nuclear norm of the consensus affinity matrix to endow it with the low-rank property.
Xu et al. \cite{esa2024xu} employed an autoencoder to extract multi-layer enhanced features.
For each layer of these enhanced features, a distinct affinity matrix was constructed per viewpoint.
The low-rank property was attained by minimizing the nuclear norm across these view-specific and layer-specific affinity matrices, culminating in a consensus affinity matrix.
In contrast, Chen et al. \cite{pr2020chen} introduced an alternative approach to establish the low-rank characteristic.
Analogous to \cite{tpami2023}, they synthesized a tensor from all view-specific affinity matrices;
however, their innovation lay in minimizing the t-SVD (tensor Singular Value Decomposition) based nuclear norm of this composite tensor.
Further extensions of sparse and low-rank feature extraction were explored in
the matrix factorization method \cite{yang1}, the graph-based method \cite{yang2}
and the multitask multiview clustering method \cite{yang3}.

\begin{table*}[htpb]
    \centering
    \caption{Some differences among some existing methods and our CL-LSR\red{.}}
    \label{somemethod}
    \resizebox{\textwidth}{!}{
    \begin{tabular}{cccccc}
        \hline
        Generalized model &local information &\makecell[c]{low-rank or sparse\\ information} &$\phi(C^1, \cdots, C^V)$ &$\varphi(C^v,C^*)$ &sources\\
        \hline
        \multirow{5}{*}{Model \eqref{generalfd}} &smoothness term on $C^v$ &- &$\sum_{v\neq w}\text{HSIC}(C^v,C^w)$ &- &\cite{cvpr2015} \\
        &- &$\ell_1$ norm on $C^v$ &$\sum_{1\leq v<w}\|C^v-C^w\|_1$ &- &\cite{neuro2015} \\
        &- &nuclear norm on $C^v$ &$\sum_{v\neq w}\|C^v-C^w\|_F^2$ &- &\cite{mlrr} \\
        &- &$\ell_1$ norm and nuclear norm on $C^v$ &$\sum_{v\neq w}\|C^v-C^w\|_F^2$ &- &\cite{pr2018} \\
        &- &\makecell[c]{tensor logarithmic Schatten-$p$ norm\\on the tensor generated from $C^v$} &- &- &\cite{tpami2023} \\
        \cmidrule{2-6}
        \multirow{7}{*}{Model \eqref{generalsd}} &smoothness term on $C^*$ &nuclear norm on $C^*$ &- &$C=C*C^*+E$ &\cite{neuro2020} \\
        &smoothness term on $C^*$ &nuclear norm on $C^*$ &- &$C^v=C^*+E^v$ &\cite{jsmc} \\
        &\makecell[c]{rank constraint on Laplacian\\matrix of $C^v$} &- &- &$\|F^v-F^*\|_F^2$ &\cite{Zhang2022} \\
        &\makecell[c]{rank constraints on Laplacian\\matrix on $C^v$ and $C^*$} &- &- &rank($L_{C^v}$)=rank($L_{C^*}$) &\cite{subspaceKinds} \\
        &- &nuclear norm on $C^v_i$ and $C^*$ &- &$\|C^v_i-C^*\|_F^2$ &\cite{esa2024xu}\\
        &smoothness term on $C^v$ &\makecell[c]{t-SVD-based tensor nuclear norm\\on the tensor generated from $C^v$} &- &$\|C^v-C^*\|_F^2$ &\cite{pr2020chen} \\
        &cardinality ($\ell_0$ norm) constraints  on $C^v$ &low-rank constraint on $C^*$ &- &$\|C^v-C^*\|_F^2$ &CL-LSR \\
        \hline
    \end{tabular}
    }
\end{table*}

For comparison, we summarize the above-mentioned methods and our proposed method in Table \ref{somemethod},
where the symbol ``-" indicates that the methods do not have relevant designs.
From Table \ref{somemethod},
one observes that the existing methods have made great progress in learning local information and
obtaining low-rank or sparse affinity matrices.
However, several shortcomings persist in the current methods.
\textit{First of all}, regardless of whether local information is captured through the smoothness term or the rank constraint, the reliance on Laplacian matrices is prevalent.
Notably, neither the smoothness term nor the rank constraint is directly applied to the affinity matrix. This indirect approach may lead to a reduction in the effectiveness of leveraging local information for clustering.
\textit{Secondly}, the pursuit of the low-rank property in the affinity matrix often involves the use of relaxed norms of the rank function.
Unfortunately, these relaxed norms may not precisely replicate the performance characteristics of the true rank function.
As a result, the achieved low-rank approximation may deviate from the optimal solution expected in an ideal scenario.
\textit{Last but not least}, augmented Lagrange method (ALM)-based alternating minimization algorithms are abused
in solving the clustering models.
However, these algorithms cannot guarantee convergence when meeting the nonconvex nonsmooth models \cite{chen2016}.
Although they may achieve satisfactory clustering results on a small number of datasets,
the instability of the algorithms always hinders the generalization ability in practical applications.

\begin{figure}[H]
    \centering
    \includegraphics[scale=0.35]{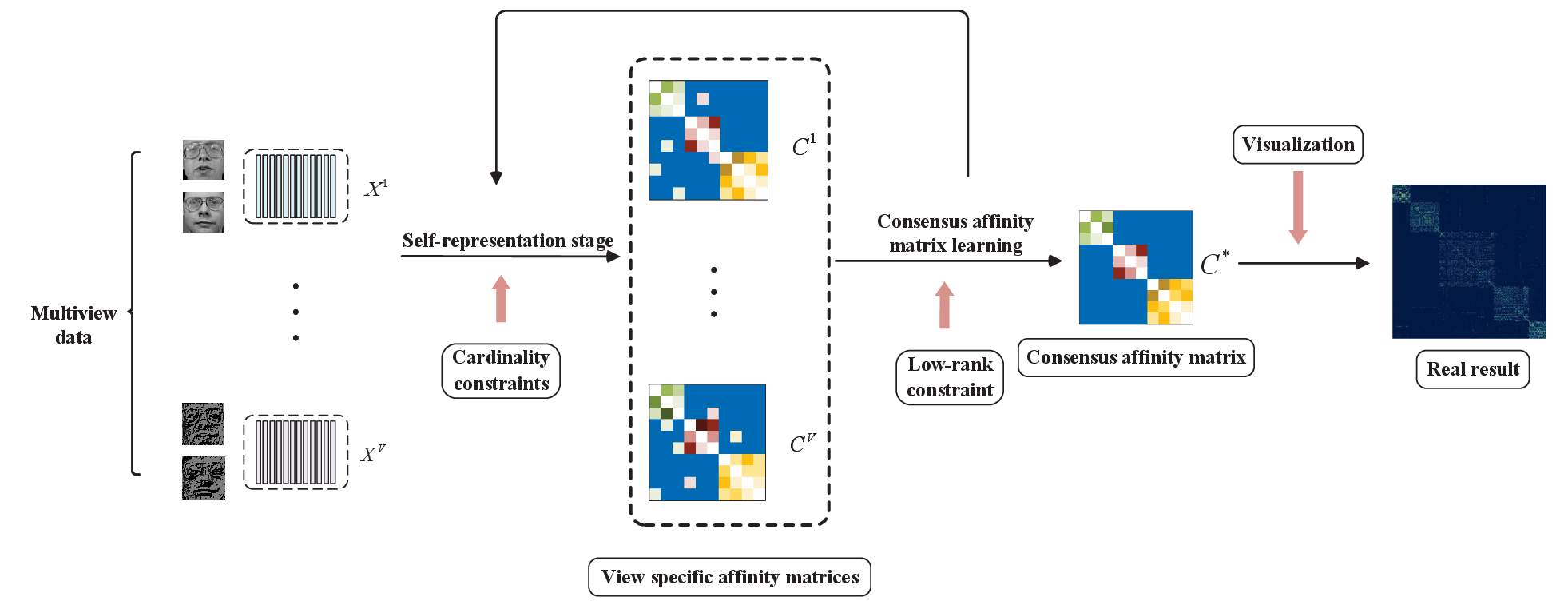}%
    \caption{CL-LSR framework. In the self-representation stage,
    the cardinality constraints excavate reliable local and global structure information under each view;
    Simultaneously, a low-rank constraint makes the consensus affinity matrix with a clear global structure.
    The synergy of two types of sparse constraints can boost each other in our CL-LSR
    for obtaining a good `real result', which is a visualization example on BBCsport dataset.
    See Subsection 4.2 for more details on the `real result'.}
    \label{framework12}
\end{figure}

To address the above, we propose a joint sparse self-representation learning
for the subspace clustering by a novel boosted model,
termed \textit{Cardinality-constrained Low-rank Least Squares Regression} (CL-LSR).
As shown in Fig. \ref{framework12},
in the self-representation stage, the cardinality constraints directly applied on the affinity matrices can restrict the number of samples to discover the reliable local information.
In the consensus information learning stage, a low-rank consensus affinity matrix is obtained with
a directly rank constraint than using some relaxed norms to approach the low-rank property.
However, the resulting clustering model is nonconvex and nonsmooth,
resulting in an attendant challenge that the existing alternating minimization methods
cannot be directly applied to our proposed model.
Therefore, we present a computationally inexpensive algorithm
called the alternating quadratic penalty method, and make a rigorous convergence analysis.

This paper has two main contributions.
\bi
    \item[(i)]
    In terms of modeling, we propose a novel cardinality-constrained low-rank least squares regression (CL-LSR) model for multiview subspace clustering.
    We inherit both local and complementary structural information by modeling the synergy of the cardinality and low-rank constraints.

    \item[(ii)] In terms of solving, we develop an efficient
    alternating quadratic penalty method,
    where each subproblem is solved iteratively by explicit solutions.
    Moreover, we make an in-depth convergence analysis, and obtain favorable performance
    on six multiview datasets, compared to eight state-of-the-art clustering methods.
\ei

The organization of this paper is following.
Section 2 gives some necessary mathematical preliminaries.
Section 3 details the proposed CL-LSR model, the corresponding algorithm with global convergence.
Section 4 displays the related experimental results.
Section 5 presents some concluding remarks.

\section{Mathematical Preliminaries}

In this section, a brief review of certain mathematical principles and essential sparse concepts is presented.

In subspace clustering, the number of non-zero elements in columns (or rows) of the affinity matrix significantly impacts clustering results.
If a column (or row) contains too many non-zero elements, it becomes difficult to accurately capture the global structure.
Thus, incorporating cardinality constraints is crucial for subspace clustering algorithms.
For a given vector $x \in \mathbb{R}^n$, the $\ell_0$ norm represents the number of non-zero elements in $x$.
A cardinality constraint on $x$ denotes the number of non-zero elements cannot exceed
a given threshold value, formulated as
\begin{equation}
    \label{cardi}
    \|x\|_0\leq k,
\end{equation}
where $k$ is the given integer.
Unfortunately, since the cardinality-constrained clustering model is generally a NP-hard problem,
the convex or non-convex approximation constraint, such as $\ell_1$ or $\ell_{1/2}$ norm,
is developed to relax the cardinality constraint.
These relax models are very popular because they are solved easily by using existing algorithms,
such as the soft and half thresholding iterative algorithms \cite{tibsh96, xu2012}.
Distinct from other methods, in this paper,
we develop a new algorithm for solving the cardinality-constrained clustering model directly.

The following is a simple introduction for the penalty method.
Consider such a general constrained minimization problem
\begin{equation}
    \label{pm}
    \begin{aligned}
        \min\quad &f(x) \\
        \text{s.t.}\quad & c(x)=0,
    \end{aligned}
\end{equation}
where $f: \mathbb{R}^n \rightarrow \mathbb{R}$ and $c: \mathbb{R}^n \rightarrow \mathbb{R}^m$ denote
objective and constraint functions, respectively.
Then, the penalty method is to reformulate \eqref{pm} as an unconstrained model \eqref{pmp} and solve it
by utilizing a penalty function $\phi(\cdot)$, written as
\begin{equation}
    \label{pmp}
    \min\quad f(x)+\phi(c(x)).
\end{equation}
The penalty function $\phi(\cdot)$ has two popular mathematical forms.
The first one is the augment Lagrange penalty function:
\begin{equation}
    \label{pmag}
    \phi(c(x))=c(x)^T b + \frac{\mu}{2}\|c(x)\|_2^2,
\end{equation}
where $b$ is the Lagrange multiplier and $\mu$ is the given penalty coefficient.
Among the clustering methods, the popular alternating direction algorithms are designed
based on \eqref{pmag}.
However, for most nonconvex nonsmooth problems,
these algorithms lack global convergence \cite{chen2016},
and thus do not have good generalization ability.
The second one is the quadratic penalty function:
\begin{equation}
    \label{pmqp}
    \phi(c(x))=\frac{\sigma}{2}\|c(x)\|_2^2,
\end{equation}
where $\sigma$ is the penalty parameter.
In this paper, in order to handle the bi-sparse (i.e., low-rank and cardinality) constraints simultaneously, we develop a new alternating quadratic penalty method with global convergence based on \eqref{pmqp}.
More details about the proposed method will be discussed in the next section.

\section{Cardinality-Constrained Low-rank Least Squares Regression}
In this section, we establish a cardinality-constrained low-rank least squares
regression (CL-LSR) model for multiview subspace clustering,
and propose an efficient alternating quadratic penalty (AQP) method
with global convergence.

\subsection{The proposed subspace clustering model}
As the work \cite{Liu2010} pointed out, a sparse affinity matrix is beneficial for subspace clustering.
A cardinality-constrained least squares
regression (C-LSR) model is developed basing on \cite{LSR}, which can be expressed as
\beq\label{c-lsr}
\min\limits_{C\in \Psi}~ \dfrac{1}{2} \| X-XC\|_F^2 + \lambda\| C\|_F^2,
\eeq
where
$\Psi :=\left\{C \in \mathbb{R}^{n \times n} :
\text{diag}(C)=\bm{0}, C \geq 0, \| C_i\|_0\leq k_1, i \in [n] \right\}$,
$k_1 \in \mathbb{Z}_+$ is the upper bound on the number of non-zero elements in column $i$ of the matrix $C$, satisfying $k_1 \ll n$.
The Frobenius norm is chosen as the measure function for self-representation and the regularization term.
Although the Frobenius norm is sensitive to noise and outliers, the cardinality constraints on $C$ could
guarantee the quality of the learned affinity matrix $C$.
By limiting the number of samples used in the reconstruction,
it contributes to finding the primary structure information and generating a sparse structure.
In addition, since $C$ stores the similarity information of data,
implicitly requiring non-negativity, we impose $C \geq 0$.
In the absence of non-negative constraints,
the C-LSR model degenerates into Yang et al. \cite{ksparseLSR}.
Clearly, model \eqref{c-lsr} is a single-view method.

Further, we develop a cardinality-constrained low-rank least squares
regression (CL-LSR) model of multiview subspace clustering based on \eqref{c-lsr}
by introducing the consensus affinity matrix $C^*$
and simultaneously optimizing the loss functions across all $V$ views.
This model can be formulated mathematically as
\beq\label{cl-lsr}
\min\limits_{C^v\in \Psi^v, C^*\in \Omega} \left\{
\sum\limits_{v=1}^V \dfrac{1}{2} \| X^v-X^vC^v\|_F^2+\lambda\| C^v\|_F^2:
C^v = C^*, v \in [V]  \right\},
\eeq
where
$\Psi^v :=\left\{C^v \in \mathbb{R}^{n \times n} :
\text{diag}(C^v)=\bm{0}, C^v \geq 0, \| C^v_i\|_0\leq k_1, i \in [n] \right\},\forall v \in [V]$;
$\Omega := \left\{C^* \in \mathbb{R}^{n \times n}:\text{rank}(C^*)\leq k_2\right\}$
with $k_2 \in \mathbb{Z}_+$ and $k_2 \ll n$. 
In problem \eqref{cl-lsr}, to obtain a clear structure of $C^*$, we impose a low-rank constraint on $C^*$,
which requires that the relationships of all samples
should be reconstructed on $k_2$ basis vectors in the consensus matrix $C^*$.
Under the cardinality and low-rank constraints,
the CL-LSR model is nonconvex and nonsmooth.
As mentioned above, it is not suitable for the alternating direction algorithms.
To solve \eqref{cl-lsr}, we focus on the quadratic penalty method
and attempt to develop a new penalty algorithm with global convergence.

Notice that, in practical, it is impossible to satisfy $C^v = C^*$ for each $ v \in [V]$.
It is more reasonable to solve an approximate solution $(\{C^v\}_{v=1}^V, C^*)$ satisfying $\| C^v-C^*\|_F \leq \epsilon$,
where $\epsilon > 0$ is an error tolerance.
It denotes that the $v$-th view affinity matrix $C^v$ is similar to the consensus affinity matrix $C^*$ for any $v \in [V]$.
Therefore, we make the following two mild assumptions,
and attempt to solve an $\epsilon$-approximation solution.

\begin{assumption}\label{assum1}
Problem \eqref{cl-lsr} is $\epsilon$-feasible, i.e., there exists at least a solution, denoted by $\left(\{C^v_{feas}\}_{v=1}^V, C^*_{feas}\right)$,
satisfying all constraints in \eqref{cl-lsr} except $C^v = C^*, \forall~ v \in [V]$.
But it satisfies that $\|C^v_{feas} - C^*_{feas}\|_F^2 \leq \epsilon^v$ for any $v \in [V]$ and a given $\epsilon^v > 0$.
Specially, if $\epsilon^v =0,~ \forall~v \in [V]$, then problem \eqref{cl-lsr} is feasible.
\end{assumption}

\begin{assumption}\label{assum2}
    Suppose the input data matrices $\{X^v\}_{v=1}^V$ are all originally bounded,
    i.e., there exists a $M > 0$ subject to $|X^v_{i,j}| \leq M$ for any $i \in [m]$
    and $ j \in [n]$.
    Moreover, there always exists a self-representation structure $\{C^v\}_{v=1}^V$
    for all the views, satisfying $\sum\limits_{v=1}^V \|X^v - X^v C^v \| < +\infty$.
\end{assumption}

The above assumptions will be used to design the AQP method
with nice convergence properties.
It can be dropped, but the theoretical convergence of the corresponding AQP method may become weaker.
We shall also mention that, for numerous real applications,
Assumptions \ref{assum1}--\ref{assum2} are readily satisfied or can be observed from
the physical background of problems.

\subsection{An Alternating Quadratic Penalty Method for CL-LSR}
Next, we propose an Alternating Quadratic Penalty (AQP) method to solve the CL-LSR model \eqref{cl-lsr}, in which a Block Coordinate Descend (BCD) method \cite{Lu2013}
is adjusted to solve each subproblem.

First, we can get the quadratic penalty function of \eqref{cl-lsr} with respect to the penalty factor $\sigma>0$. That is, for any $C^v\in \Psi^v,~ C^*\in \Omega$,
\[
 q_\sigma(\{C^v\}_{v=1}^V, C^*) := \sum_{v=1}^V \dfrac{1}{2} \| X^v-X^vC^v\|_F^2+\lambda\| C^v\|_F^2 + \dfrac{\sigma}{2}\| C^v-C^*\|_F^2.
\]
Then, the AQP method for the CL-LSR model can be described below.\\

\noindent
{\bf AQP Method for Model \eqref{cl-lsr}}
\vspace{0.05in}\\
Given a positive decreasing
sequence $\{\epsilon_k\}$, initial value $\sigma_0>0$ and $\rho>1$,
an arbitrarily selected constant $C^*_0 \in \Omega$ and $k=0$.
\begin{enumerate}
\renewcommand{\labelenumi}{\bf \theenumi$^\circ$}
\item
Let $l=0$ and apply the following BCD method to obtain an approximate optimal solution $(\{C^v_k\}_{v=1}^V, C^*_k)\in \prod\limits_{v=1}^V \Psi^v \times \Omega$,
\begin{equation}\label{BCD}
\min~ \{q_{\sigma_k}(\{C^v\}_{v=1}^V, C^*),  ~\forall~ C^v\in \Psi^v,~ C^*\in \Omega\}.
\end{equation}
\begin{enumerate}[a)]

\item Obtain an optimal solution by solving the following subproblem:
\beq\label{subproblem1}
\{C^v_{k,l+1}\}_{v=1}^V \in \Arg \min~ q_{\sigma_k}(\{C^v\}_{v=1}^V ,C^*_{k,l}),
~ \forall~ C^v \in \Psi^v, ~v \in [V];
\eeq

\item Solve an optimal solution
\beq\label{subproblem2}
C^*_{k,l+1} \in \Arg \min\limits_{C^*\in \Omega}q_{\sigma_k}(\{C^v_{k,l+1}\}_{v=1}^V,C^*);
\eeq

\item Set $(\{C^v_k\}_{v=1}^V,C^*_k):=(\{C^v_{k,l+1}\}_{v=1}^V,C^*_{k,l+1})$.
If the condition
\begin{equation}\label{stop1}
\max \left\{
\| \mathcal{P}_{\Psi^v}\left(C^v_k-\bigtriangledown_{C^v}
q_{\sigma_k}(\{C^v\}_{v=1}^V,C^*_k)\right)-C^v_k \|_F^2 \right\}_{v=1}^V \leq \varepsilon_k
\end{equation}
is satisfied then go to {\bf 2$^\circ$}. Otherwise, set $l\leftarrow l+1$, and go to a).
\end{enumerate}

\item Let $\sigma_{k+1}:=\rho\sigma_k$, and set $C^*_{k+1,0}:=C^*_k$.
\item Set $k\leftarrow k+1$, then go to {\bf 1$^\circ$}.
\end{enumerate}

In the AQP method,
\eqref{stop1} is used to guarantee the global convergence.
Since the sequence $\{q_{\sigma_k}(\{C^v\}_{v=1}^V,C^*_{k,l})\}$ is nonincreasing,
we can terminate above method based on the progress of
$\{q_{\sigma_k}(\{C^v_{k,l}\}_{v=1}^V,C^*_{k,l})\}$.
Therefore, the internal stop criterion \eqref{stop1} can be replaced by
\begin{equation}\label{stop2}
\max\{\frac{\| C^v_{k,l}-C^v_{k,l-1}\|_F}{\max(\|
C^v_{k,l}\|_F,1)},\frac{\| C^*_{k,l}-C^*_{k,l-1}\|_F}{\max(\|
C^*_{k,l}\|_F,1)}\}_{v=1}^V\leq \varepsilon_I,
\end{equation}
where $\varepsilon_I>0$ is a given internal tolerance.
The following condition
\begin{equation}\label{stop-out}
\max\{\|C^v_k-C^*_k\|_F\}_{v=1}^V \leq \varepsilon_O
\end{equation}
is used as an external stop criterion, where $\varepsilon_O>0$ is a given external tolerance.

\subsection{The solutions for subproblems}

The AQP method focuses on solving subproblems \eqref{subproblem1} and \eqref{subproblem2} alternately.
We next establish the efficient explicit iterations for the two subproblems. \\

1) {\bf An efficient explicit iterations for \eqref{subproblem1}}

Subproblem \eqref{subproblem1} can be decomposed into
$V$ sub-blocks for parallel processing.
Under the $v$-th view, the resulting sub-block is formulated as
\begin{equation}\label{subproblem1o2}
\min\limits_{C^v\in \Psi^v}~ \dfrac{1}{2}\|X^v-X^vC^v\|_F^2+\lambda\|C^v\|_F^2+\dfrac{\sigma_k}{2}\|C^v-C^*_{k,l}\|_F^2.
\end{equation}
Further, we split \eqref{subproblem1o2} by the columns of $C^v$,
and simultaneously handle $diag(C^v)=\mathbf{0}$ by deleting the corresponding elements.
Then, we obtain such a split form:
\begin{equation}
    \label{subproblem1o3}
    \ba{cl}
        \min\limits_{\bar{C^v_i}\in \mathbb{R}^{n-1}}~ & \sum\limits_{i=1}^n ~\dfrac{1}{2}\| X^v_i-X^v_{-i}\bar{C^v_i}\|_2^2+\lambda\|\bar{C^v_i}\|_2^2+\dfrac{\sigma_k}{2}\|\bar{C^v_i}-[\bar{C^*_{k,l}}]_{i}\|_2^2 \\
        \text{s.t.}~ &\|\bar{C^v_i}\|_0\leq k_1, \bar{C^v_i}\geq 0, i\in[n],
    \ea
\end{equation}
where $X^v_i\in \mathbb{R}^{n \times 1}$ is the $i$-th column of $X^v$;
$X^v_{-i}\in \mathbb{R}^{n\times(n-1)}$ is the submatrix of $X^v$
which deletes the $i$-th column;
$\bar {X^v_i} \in \mathbb{R}^{(n-1)\times 1}$ is a vector obtained by removing
the $i$-th element in $X^v_i$.
Let $x=\bar{C^v_i}$, $b=X^v_i$, $A=X^v_{-i}$ and $c=[\bar{C^*_{k,l}}]_{i}$,
then \eqref{subproblem1o3} can be rewritten as
\begin{equation}\label{subproblem1o5}
    \min\limits_{x \in \mathbb{R}^{n-1}}~ \left\{ \|\tilde{A}x-\tilde{b}\|_2^2:~\|x\|_0\leq k_1, x \geq 0 \right\},
\end{equation}
where $\tilde{A}=\left[
\dfrac{1}{\sqrt{2}} A,  \sqrt{\lambda}\mathrm{I}, \sqrt{\dfrac{\sigma_k}{2}} \mathrm{I} \right]^T$, $\tilde{b}= \left[ \dfrac{1}{\sqrt{2}}b, \mathbf{0}, \sqrt{\dfrac{\sigma_k}{2}}c \right]^T$.
It is clear to see that \eqref{subproblem1o5} is a special case of the following optimization
\beq \label{sparse-prob}
\min\limits_{y \in \Delta_{k_1}^+} f(y),
\eeq
where $\Delta_{k_1}^+:=\left\{y \in \mathbb{R}^{n}: \|y\|_0 \leq k_1, y \geq 0 \right\}$,
$f: \mathbb{R}^n \to \mathbb{R}$ is Lipschitz continuously differentiable,
i.e., there is a constant $L_f > 0$ such that
\[
\left\| {\nabla f(y) - \nabla f(z)} \right\|_2 \le L_f\left\| {y - z} \right\|_2, \quad \forall y, z \in \mathbb{R}^n.
\]

Next, we apply a nonmonotone projected gradient (NPG) method for \eqref{sparse-prob}.
Apparently, the NPG method below is suitable for solving \eqref{subproblem1o5}, or \eqref{subproblem1} in parallel. \\

\noindent
{\bf Nonmonotone projected gradient (NPG) method for \eqref{sparse-prob}}

Let $0< L_{\min} < L_{\max}$, $\tau>1$, $c>0, M \in \mathbb{N}_+$ be given. Choose an
arbitrary $y^0 \in \Delta_{k_1}^+$ and set $k=0$.
\begin{enumerate}
\renewcommand{\labelenumi}{\bf \theenumi$^\circ$}
\item Choose $L^0_k \in [L_{\min}, L_{\max}]$ arbitrarily. Set $L_k = L^0_k$.
\bi
\item[a)] Solve the subproblem
\beq \label{subprob}
y^{k+1} \in \Arg\min\limits_{y \in \Delta_{k_1}^+} \left\{\nabla f(y^k)^T (y-y^k) + \frac{L_k}{2} \|y-y^k\|_2^2\right\}.
\eeq
\item[b)] If
\beq \label{descent}
f(y^{k+1}) \le \max\limits_{\max\{k-M, 0\} \le i \le k} f(y^i) - \frac{c}{2} \|y^{k+1}-y^k\|_2^2
\eeq
is satisfied, then go to step {\bf 2$^\circ$}.
\item[c)] Set $L_k \leftarrow \tau L_k$ and go to step a).
\ei
\item Set $k \leftarrow k+1$ and go to step {\bf 1$^\circ$}.
\end{enumerate}

In the NPG method, we suggest that a smart choice of $L^0_k$ is the BB-step proposed by Barzilai and Borwe in \cite{bb-step88} (see also \cite{birgin2000}):
\[
L^0_k  = \max \left\{ L_{\min } ,\min \left\{ L_{\max } ,\frac{(y^k  - y^{k - 1})^T \nabla f(x^k)-\nabla f(x^{k - 1})}{\|y^k  - y^{k - 1}\|^2}\right\} \right\}.
\]
The performance of the method mainly depends on the calculated cost of the projection \eqref{subprob}.
Fortunately, it follows from Lemma 2.1 in \cite{Zhao2015adaptive} that
we obtain the closed-form solution of \eqref{subprob} as follows:
\beq \label{closed-form solution1}
y^*_i = \left\{
\ba{cl}
\max\{0, z^k_i\}, & i \in \mathcal{I}_{k_1}(z^k), \\[0.2cm]
0, & \text{otherwise},
\ea
\right.
\eeq
where $z^k = y^k - \dfrac{1}{L_k} \nabla f(y^k)$ and
$\mathcal{I}_{k_1}(z^k)$ is the indices of $k_1$ largest entries of $z^k$. \\

\begin{itemize}
        \item [2)] \bf{The closed-form solution of \eqref{subproblem2}}
\end{itemize}

For subproblem \eqref{subproblem2}, one observes that it is equivalent to
\[
\ba{cl}
& \min ~ \sum\limits_{v=1}^V \dfrac{1}{V}\|C^v_{k,l+1}-C^*\|_F^2 \\
\Leftrightarrow  & \min~\sum\limits_{v=1}^V \dfrac{1}{V} tr({C^v_{k,l+1}}^TC^v_{k,l+1})-2tr({C^*}^T\sum_{v=1}^V\frac{1}{V}C^v_{k,l+1}) + tr({C^*}^TC^*) \\
\Leftrightarrow  & \min ~tr({C^*}^TC^*)-2tr\left(\left(\sum_{v=1}^V\frac{1}{V}C^v_{k,l+1}\right)^T C^*\right) \\
\Leftrightarrow  & \min~\|C^*-W\|_F^2 ~ \text{with}~W=\frac{1}{V} \sum_{v=1}^V C^v_{k,l+1}.
\ea
\]
Therefore, in the following, we only need to solve
\begin{equation}\label{subproblem2o3}
        \min\limits_{C^*\in \Omega}~  \|C^*-W\|_F^2.
\end{equation}
Without loss of generality,
define the singular value decomposition (SVD) of $W$ by $W = U_W \Sigma_W V_W^T$,
where $\Sigma_W = \mathcal{D}\left(\sigma_1, \sigma_2, ...,\sigma_n \right)$
is a diagonal matrix satisfying $\sigma_1 \geq \sigma_2 \geq... \geq\sigma_n \geq 0$.
It then follows from the Eckart and Young's Theorem \cite{Eckart1936}
that the best rank-$k_2$ approximation to the matrix $W$ is
its truncated SVD with $k_2$-terms.
That is, \eqref{subproblem2o3} or \eqref{subproblem2}
admits the following closed-form solution
\begin{equation}\label{eqcstar2}
    C^* = U_W \Sigma_W^* V_W^T,
\end{equation}
where $\Sigma_W^* = \mathcal{D}(\sigma_1,\sigma_2,...,\sigma_{k_2},0,...,0)$.

\subsection{Convergence analysis}

Next, we make the convergence analysis for our proposed AQP method.

\begin{theorem}\label{th3-1}
Let $\{(\{C^v_{k,l}\}_{v=1}^V, C^*_{k,l})\}_{l=1}^{\infty}$
be the sequence generated by applying the BCD method to problem
\eqref{BCD} for a given $\varepsilon_k>0$. There holds:
\bi
\item[(i)] $\{q_{\sigma_k}(\{C^v_{k,l}\}_{v=1}^V, C^*_{k,l})\}$ is a nonincreasing sequence with respect to $l>1$;

\item[(ii)]~Any accumulation point of $\{(\{C^v_{k,l}\}_{v=1}^V, C^*_{k,l})\}_{l=1}^{\infty}$ is a block coordinate minimizer of problem \eqref{BCD};

\item[(iii)]~There exists a $\bar{l}>0$, such that for all $l>\bar{l}$,
\begin{equation}
\max \{
\| \mathcal{P}_{\Psi^v}(C^v_{k,l}-\bigtriangledown_{C^v}
q_{\sigma_k}(\{C^v_{k,l}\}_{v=1}^V,C^*_{k,l}))-C^v_{k,l} \|_F^2 \}_{v=1}^V \leq \varepsilon_k.
\end{equation}
\ei
\end{theorem}
\begin{proof}
(i)~Since
\begin{equation}\label{eq3-4-1}
q_{\sigma_k}(\{C^v_{k,l+1}\}_{v=1}^V,C^*_{k,l})\leq q_{\sigma_k}(\{C^v_{feas}\}_{v=1}^V,C^*_{k,l}),~
\text{for}~\forall C^v_{feas} \in \Psi^v, v \in [V],
\end{equation}
\begin{equation}\label{eq3-4-2}
q_{\sigma_k}(\{C^v_{k,l}\}_{v=1}^V,C^*_{k,l})\leq q_{\sigma_k}(\{C^v_{k,l}\}_{v=1}^V,C^*_{feas}),~
\text{for}~\forall~C^*_{feas} \in \Omega,
\end{equation}
then for $\forall~l>1$, we have
\begin{equation}\label{eq3-4-3}
q_{\sigma_k}(\{C^v_{k,l+1}\}_{v=1}^V,C^*_{k,l+1})\leq
q_{\sigma_k}(\{C^v_{k,l+1}\}_{v=1}^V,C^*_{k,l})
\leq q_{\sigma_k}(\{C^v_{k,l}\}_{v=1}^V,C^*_{k,l}).
\end{equation}
Therefore, the statement (i) holds true.

(ii) It follows from Assumption \ref{assum2} that
the sequence $\{C^v_{k,l}\}_{v=1}^V$ is bounded for any $v \in [V]$.
Together with \eqref{eqcstar2}, we have that
the sequence $\{C^*_{k,l}\}$ is also bounded when
$\{C^v_{k,l}\}_{v=1}^V$ is bounded.
Therefore, there exists at least a subset $L$ of
$\{1,2,\cdots,\}$ and $(\{C^v_k\}_{v=1}^V, C^*_k)$ satisfying
\begin{equation}\label{eq3-4-4}
\lim_{l\in L\rightarrow \infty}(\{C^v_{k,l}\}_{v=1}^V,C^*_{k,l})=(\{C^v_k\}_{v=1}^V, C^*_k).
\end{equation}
Together with the statement (i) and the
continuity of $q_{\sigma_k}(\cdot,\cdot)$,
we know that the sequence $\{q_{\sigma_k}(\{C^v_{k,l}\}_{v=1}^V, C^*_{k,l})\}$ is bounded. Then,
$\lim\limits_{l \rightarrow \infty} q_{\sigma_k}(\{C^v_{k,l}\}_{v=1}^V, C^*_{k,l})$ exists
in view that it is a bounded and nonincreasing sequence.
Together with \eqref{eq3-4-3}, \eqref{eq3-4-4}
and continuity of $q_{\sigma_k}(\cdot,\cdot)$,
we have
\[
\lim_{l\rightarrow \infty} q_{\sigma_k}(\{C^v_{k,l+1}\}_{v=1}^V, C^*_{k,l})
= \lim_{l\rightarrow \infty} q_{\sigma_k}(\{C^v_{k,l}\}_{v=1}^V, C^*_{k,l}) =q_{\sigma_k}(\{C^v_k\}_{v=1}^V, C^*_k).
\]
Taking limits on both sides of \eqref{eq3-4-1} and \eqref{eq3-4-2} as $l \in L
\rightarrow\infty$,
we have
\begin{equation}\label{eq3-4-5}
q_{\sigma_k}(\{C^v_k\}_{v=1}^V, C^*_k) \leq q_{\sigma_k}(\{C^v_{feas}\}_{v=1}^V, C^*_k)~ \text{for}~\forall~C^v_{feas} \in \Psi^v,~v \in [V];
\end{equation}
\begin{equation}\label{eq3-4-6}
q_{\sigma_k}(\{C^v_k\}_{v=1}^V, C^*_k) \leq
q_{\sigma_k}(\{C^v_k\}_{v=1}^V, C^*_{feas})~\text{for}~\forall~C^*_{feas} \in \Omega.
\end{equation}
Together with \eqref{eq3-4-5} and \eqref{eq3-4-6}, we obtain the statement (ii).

(iii) Together with \eqref{eq3-4-5} and Theorem 4.2 in \cite{Lu2013},
the statement (iii) holds.
\end{proof}

\begin{theorem}\label{th3-2}
Let $(\{C^v_k\}_{v=1}^V, C^*_k)$ be the sequence generated by the AQP
method applied to problem \eqref{cl-lsr}. Then the following statements hold.
\bi
\item[(i)]~The sequence $\{(\{C^v_k\}_{v=1}^V, C^*_k)\}$ is bounded;

\item[(ii)]~Let $(\{C^v\}_{v=1}^V, C^*)$ be an accumulation point of the sequence $\{(\{C^v_k\}_{v=1}^V, C^*_k)\}$.
     Under Assumptions \ref{assum1} and \ref{assum2}, there holds that for a sufficiently small $\epsilon > 0$,
    \[
    \|C^* - \frac{1}{V} \sum\limits_{v=1}^V C^v\|_F^2 \leq \min \{\frac{1}{V} \sum\limits_{v=1}^V \epsilon^v, \epsilon_O^2\} + \epsilon.
    \]
\ei
\end{theorem}

\begin{proof}
(i) According to Assumption \ref{assum2},
we know that the sequence $\{C^v_k\}_{v=1}^V$ is bounded.
In view of the expression of $C^*_k$,
the sequence $\{C^*_k\}$ is also bounded.
Therefore, $\{(\{C^v_k\}_{v=1}^V, C^*_k)\}$ is a bounded sequence.

(ii)
According to the bounded sequence $\{(\{C^v_k\}_{v=1}^V, C^*_k)\}$,
there exists at least a subset $\mathcal{K}$ of $\{1,2,\dots\}$ and $(\{C^v\}_{v=1}^V, C^*) $ such that
\[
\lim_{k\in \mathcal{K}\rightarrow\infty}(\{C^v_k\}_{v=1}^V, C^*_k)=(\{C^v\}_{v=1}^V, C^*)
\]
holds. Then, followed by the iteration of the AQP method, we have
\[
C^*_{k,l} \in \Arg\min\limits_{C^*\in \Omega}~  \|C^*-W\|_F^2,
\]
where $W = \frac{1}{V} \sum_{v=1}^V  C^v_{k,l}$.
For the above subsets $\mathcal{L}$ and $\mathcal{K}$ of $\{1,2,\dots\}$,
\beq\label{eq3-4-7}
\ba{lll}
& & \lim\limits_{k\in \mathcal{K}\rightarrow\infty} \lim\limits_{l \in \mathcal{L} \rightarrow\infty}
\|C^*_{k,l} - \frac{1}{V} \sum_{v=1}^V  C^v_{k,l}\|_F^2 \\
& = &  \lim_{k\in \mathcal{K}\rightarrow\infty}
\|C^*_k - \frac{1}{V} \sum\limits_{v=1}^V C^v_k\|_F^2 \\
& = &  \|C^* - \frac{1}{V} \sum\limits_{v=1}^V C^v\|_F^2.
\ea
\eeq
Then, for a sufficiently small $\epsilon > 0$, we have
\beqas
\|C^* - \dfrac{1}{V} \sum\limits_{v=1}^V C^v\|_F^2  & \leq &  \|C^*_{k,l} - \frac{1}{V} \sum_{v=1}^V  C^v_{k,l}\|_F^2 + \epsilon \\
& \leq & \frac{1}{V} \sum_{v=1}^V \|C^*_{k,l} - C^v_{k,l}\|_F^2 + \epsilon \\
& \leq & \frac{1}{V} \sum_{v=1}^V \epsilon^v +\epsilon,
\eeqas
where the first inequality is obtained by \eqref{eq3-4-1} and the sign-preserving theorem of limit;
the last inequality is followed by Assumption \ref{assum2} and the optimality of $(\{C^v_{k,l}\}_{v=1}^V, C^*_{k,l})$.
On the other hand, together with \eqref{stop-out} and \eqref{eq3-4-2}, we have
\beqas
\|C^* - \dfrac{1}{V} \sum\limits_{v=1}^V C^v\|_F^2
& \leq & \frac{1}{V} \sum_{v=1}^V \|C^*_{k} - C^v_{k}\|_F^2 + \epsilon \\
& \leq  & \max\{\|C^*_k-C^v_k\|_F^2\}_{v=1}^V + \epsilon \\
& \leq &  \epsilon_O^2 + \epsilon.
\eeqas
In summary, the statement (ii) holds.
\end{proof}

\subsection{Computation complexity}

The computational complexity of the AQP method for our CL-LSR is comprised of two primary components:
1) solving subproblem \eqref{subproblem1}; 2) addressing subproblem \eqref{subproblem2}.

Regarding subproblem \eqref{subproblem1}, the update rule is governed by column operations.
Specifically, when updating a single column of matrix $C^v$,
the primary computational burden is vector multiplication,
which incurs a complexity of $O(t_{i} n)$,
where $t_{i}$ denotes the number of iterations in the NPG method.
Consequently, the aggregate complexity for updating all view-specific affinity matrices
$C^v~(v\in[V])$ amounts to $O(Vt_{i}n^2)$, with $V$ representing the number of views.
In addressing subproblem \eqref{subproblem2}, a closed-form solution can be obtained through singular value decomposition (SVD).
The standard complexity associated with this step is $O(n^3)$.
Nonetheless, drawing from the work presented in \cite{linearsvd},
should a linear time SVD algorithm be employed,
the complexity may be mitigated to $O(n (k_2)^2 + (k_2)^3)$.

In summary, the total complexity of our AQP method is $O(t(V t_i n^2 + (k_2)^2 n + (k_2)^3))$,
where $t$ is the total iteration number of the AQP method.
Since $V$, $t$, $t_i$ and $k_2$ are much smaller than $n$ generally,
the complexity of the AQP method could be summarized as $O(n^2)$.

\section{Numerical Results}

In this section, we conduct numerical experiments to
demonstrate the superior performance of the CL-LSR model
with the AQP method on various data sets.
All methods were coded in Matlab, and
all computations were performed on a desktop
(Intel Core i7-6700, 3.4GHz, 8GB RAM) with Matlab R2020a.

\subsection{Experimental setup}

\subsubsection{Datasets, comparators and criteria}

Six multiview datasets are tested in our experiments as follows,
of which the details are described in Table \ref{tdata1}.

\bi
\item \textbf{ORL} \cite{orl} has three views containing 400 images of 40 human faces under different conditions, where each class has 10 images.

\item\textbf{3sources}\footnote{http://mlg.ucd.ie/datasets/3sources.html}
contains news which are collected from three sources: BBC, Reuters and the Guardian. In experiments, we select 169 samples reported in all the views.

\item \textbf{BBCsport}\footnote{http://mlg.ucd.ie/datasets/segment.html.}
has two views with 544 samples, which are obtained from the BBC sport website and belong to five classes.

\item \textbf{Caltech101-7} \cite{caltech} contains 101 categories. We select samples from seven different classes to generate the dataset for experiments. The selected dataset has 1474 samples and six views.

\item \textbf{WebKB}\footnote{http://www.cs.cmu.edu/afs/cs.cmu.edu/project/theo-20/www/data/} has two views with 1051 samples, which are gathered from four university network documents.

\item \textbf{wikipedia}\footnote{http://www.svcl.ucsd.edu/projects/crossmodal/} has two views with 693 samples, which are collected form the highlighted Wikipedia's articles assortment.
\ei

\begin{table}[htpb]
	\centering
        \footnotesize
        \setlength{\abovecaptionskip}{2pt}
        \setlength{\belowcaptionskip}{2pt}
	\caption{Details for six multiview datasets\red{.}} \label{tdata1}
	\begin{tabular}{cccccc}
		\hline
		Name &Views &Clusters &Samples &Features & Remarks\\
		\hline
		ORL &$3$ &$40$ &$400$ &$496,~3304,~6750$ & face pictures\\
		3sources &$3$ &$6$ &$169$ &$3560,~3631,~3068$ &news reports\\
		BBCsport &$2$ &$5$ &$544$ &$3183,~3203$ &  BBC sport\\
		Caltech101-7 &$6$ &$7$ &$1474$ &$48,~40,~254,~1984,~512,~928$ & pictures of objects\\
		WebKB &$2$ &$2$ &$1051$ &$2949,~334$ & university network \\
		wikipedia &$2$ &$10$ &$693$ &$128,~10$ & Wiki articles \\
		\hline
	\end{tabular}
\end{table}

In the experiments, we compare our proposed method with
eight state-of-the-art clustering methods,
which are summarized in the following.

\bi
\item \textbf{MLRR} \cite{mlrr} is a multiview subspace clustering method,
considering the low-rank representations and symmetric constraints
at the same time.

\item \textbf{DCEL} \cite{Mi} projects the multiview data into a
latent embedding space, and considers the diversity and consistency of
data.

\item \textbf{FPMVS-CAG} \cite{fpmvs} contains no hyperparameters,
into which an anchor strategy is embedded.

\item \textbf{GMC} \cite{gmc} is a graph-based multiview clustering
method, where a learning method is proposed to improve the quality of the
unified graph and the subgraph under each view.

\item \textbf{LMVSC} \cite{c1} is a multiview subspace clustering
algorithm which utilizes anchors to reduce the computational burden.

\item \textbf{BMVC} \cite{bmvc} uses binary coding for multiview data
to obtain the binary representations and reduce computational burden.

\item \textbf{JSMC} \cite{jsmc} is a subspace method which focus on commonness and inconsistencise
among multiview data.

\item \textbf{UDBGL} \cite{udbgl} utilizes the anchor strategy to generate bipartite graphs with
adaptive weight learning.
\ei

For comparison, we introduce the four criteria
to measure the performance of clustering:
(i) Accurary (ACC) \cite{gnmf};
(ii) Normalized Mutual Information (NMI) \cite{gnmf};
(iii) F-measure (Fs) \cite{fs};
(iv) Adjustied Rand Index (ARI) \cite{ari}.
Notice that, for all the criteria, higher values represent better clustering performance.

\subsubsection{Initialisation}

Followed by the comparative methods such as \textbf{MLRR} \cite{mlrr},
we use the PCA method to reduce the dimension of features,
where the dimension of the reduced space is set as
$ \mathop{\min}\ \{100, m_1,m_2,...,m_V\}$.
After this pretreatment, the length of features is the same for all views.
Then, we initialise a couple of parameters $(\{C^v\}_{v=1}^V, C^*)$.
Specially, for $v \in [V]$, the initialization of $C^v$ is
followed by \cite{gnmf}:
\[
    C^v_{ij}= \left\{
    \ba{cl}
        e^{-\dfrac{\| X^v_i-X^v_j\|_2^2}{\sigma}}, & \text{if} \ X_i^v\in \kappa(X^v_j) \ \text{or} \ X^v_j\in \kappa(X^v_i), \\
        0, & \text{otherwise},
    \ea
    \right.
\]
where $\kappa(X^v_i)$ is a set including the $\kappa$ nearest
neighbors for $X_i^v$; $\kappa$ and $\sigma$ are set the same as \cite{gnmf}.
Based on the setting of $\{C^v\}_{v=1}^V$, we set $ C^*=\frac{1}{V} \sum_{v=1}^VC^v$.

For the AQP method,
the parameters $\lambda$, $\sigma$, $\rho$, $\varepsilon_I$ and $\varepsilon_O$
are set as 100, 1, 10, $10^{-4}$, and $10^{-2}$.
Moreover, we set $k_1=20$, $k_2=20k_c$, where $k_c$ is the number of clusters. 
For the NPG method,
the parameters $L_{min}$, $L_{max}$, $\tau$, $c$ and $M$ are set as
$10^{-10}$, $10^{10}$, 3, $10^{-6}$ and 5, respectively.
For the proposed CL-LSR,
we use the matrix $C^*_f$ uniformly to get final clustering results,
defined as
\[
C^*_f = \frac{[C^*]_++[{C^*}]_+^T}{2},
\]
where $[\cdot]_+=max(\cdot,0)$. 
Obviously, $C^*_f$ is a symmetric and nonegative matrix
and the spectral clustering can be carried on $C^*_f$ to get clustering results.

\subsection{Comparison results}

\subsubsection{Clustering performance}

For the aforementioned six datasets,
we compare the performance of our CL-LSR method with the other methods.
To eliminate the influence of accidents,
we run 100 times for every method.
Parameters of compared methods are set as the authors' suggestion.
The resulting means and standard deviations of four evaluation measures are reported in
Table \ref{tfor4-1-all}.

\begin{table*}[!htb]
    \centering
    \caption{Clustering results on six datasets (means(standard deviations)).}
    \label{tfor4-1-all}
    \resizebox{\linewidth}{!}{
        \begin{tabular}{ccccc|cccc}
            \hline
            method &\multicolumn{4}{c}{ORL} &\multicolumn{4}{c}{3sources} \\
             &ACC &NMI &Fs &ARI &ACC &NMI &Fs &ARI \\
            \cmidrule{2-9}

            MLRR       &$0.6670_{\pm 0.0243}$  &$0.8268_{\pm 0.0122}$ &$0.6954_{\pm 0.0243}$ &$0.5549_{\pm 0.0301}$  &$0.6450_{\pm 0.0000}$  &$0.6227_{\pm 0.0000}$ &$0.7082_{\pm 0.0000}$ &$\underline{0.5582}_{\pm 0.0000}$ \\
            DCEL       &$\underline{0.6910}_{\pm 0.0240}$  &$0.8343_{\pm 0.0157}$ &$\underline{0.7163}_{\pm 0.0218}$ &$\underline{0.5811}_{\pm 0.0321}$  &$0.3100_{\pm 0.0217}$  &$0.1137_{\pm 0.0255}$ &$0.3341_{\pm 0.0272}$ &$0.0374_{\pm 0.0149}$ \\
            FPMVS-CAG  &$0.5568_{\pm 0.0051}$  &$0.7603_{\pm 0.0022}$ &$0.4136_{\pm 0.0030}$ &$0.3972_{\pm 0.0031}$  &$0.2604_{\pm 0.0000}$  &$0.0684_{\pm 0.0000}$ &$0.2146_{\pm 0.0000}$ &$0.0101_{\pm 0.0000}$\\
            GMC        &$0.6325_{\pm 0.0000}$  &$\underline{0.8571}_{\pm 0.0000}$ &$0.6959_{\pm 0.0000}$ &$0.3367_{\pm 0.0000}$ &$0.6923_{\pm 0.0000}$  &$0.6216_{\pm 0.0000}$ &$0.7041_{\pm 0.0000}$ &$0.4431_{\pm 0.0000}$ \\
            LMVSC      &$0.6190_{\pm 0.0154}$  &$0.7985_{\pm 0.0191}$ &$0.6650_{\pm 0.0175}$ &$0.0806_{\pm 0.1612}$  &$0.3947_{\pm 0.0670}$  &$0.2654_{\pm 0.0660}$ &$0.4400_{\pm 0.0678}$ &$0.0238_{\pm 0.0477}$\\
            BMVC       &$0.5820_{\pm 0.0061}$  &$0.7615_{\pm 0.0005}$ &$0.6134_{\pm 0.0046}$ &$0.4528_{\pm 0.0008}$  &$0.4083_{\pm 0.0000}$  &$0.4017_{\pm 0.0024}$ &$0.5067_{\pm 0.0005}$ &$0.2627_{\pm 0.0024}$\\
            JSMC       &$0.6250_{\pm 0.0000}$  &$0.7850_{\pm 0.0000}$ &$0.6505_{\pm 0.0000}$ &$0.4864_{\pm 0.0000}$  &$\textbf{0.7751}_{\pm 0.0000}$  &$\textbf{0.7104}_{\pm 0.0000}$ &$\textbf{0.8062}_{\pm 0.0000}$ &$\textbf{0.6517}_{\pm 0.0000}$\\
            UDBGL      &$0.5525_{\pm 0.0000}$  &$0.7691_{\pm 0.0000}$ &$0.5994_{\pm 0.0000}$ &$0.3962_{\pm 0.0000}$  &$0.3432_{\pm 0.0000}$  &$0.0553_{\pm 0.0000}$ &$0.3703_{\pm 0.0000}$ &$0.0015_{\pm 0.0000}$\\
            CL-LSR     &$\textbf{0.7777}_{\pm 0.0194}$  &$\textbf{0.9020}_{\pm 0.0057}$ &$\textbf{0.8029}_{\pm 0.0148}$ &$\textbf{0.7020}_{\pm 0.0232}$  &$\underline{0.7000}_{\pm 0.0053}$  &$\underline{0.6850}_{\pm 0.0004}$ &$\underline{0.7254}_{\pm 0.0021}$ &$0.5347_{\pm 0.0006}$\\
            \hline
            method &\multicolumn{4}{c}{BBCsport} &\multicolumn{4}{c}{Caltech101-7} \\
            &ACC &NMI &Fs &ARI &ACC &NMI &Fs &ARI  \\
            \cmidrule{2-9}
            MLRR       &$\underline{0.8915}_{\pm 0.0000}$  &$\underline{0.8091}_{\pm 0.0000}$ &$\underline{0.8885}_{\pm 0.0000}$ &$\underline{0.8292}_{\pm 0.0000}$  &$0.3786_{\pm 0.0000}$  &$0.4948_{\pm 0.0000}$ &$0.5128_{\pm 0.0000}$ &$0.3193_{\pm 0.0000}$\\
            DCEL       &$0.3234_{\pm 0.0362}$  &$0.0618_{\pm 0.0275}$ &$0.3328_{\pm 0.0320}$ &$0.0359_{\pm 0.0179}$  &$0.2709_{\pm 0.0313}$  &$0.1110_{\pm 0.0308}$ &$0.3466_{\pm 0.0408}$ &$0.0544_{\pm 0.0247}$\\
            FPMVS-CAG  &$0.4063_{\pm 0.0000}$  &$0.1058_{\pm 0.0000}$ &$0.3043_{\pm 0.0000}$ &$0.1028_{\pm 0.0000}$  &$\underline{0.6402}_{\pm 0.0524}$  &$0.5745_{\pm 0.0138}$ &$0.6389_{\pm 0.0443}$ &$\underline{0.4866}_{\pm 0.0559}$\\
            GMC        &$0.7390_{\pm 0.0000}$  &$0.7954_{\pm 0.0000}$ &$0.7917_{\pm 0.0000}$ &$0.6009_{\pm 0.0000}$  &$\textbf{0.6920}_{\pm 0.0000}$  &$\textbf{0.6595}_{\pm 0.0000}$ &$\textbf{0.7713}_{\pm 0.0000}$ &$\textbf{0.5943}_{\pm 0.0000}$\\
            LMVSC      &$0.6088_{\pm 0.0622}$  &$0.4045_{\pm 0.0636}$ &$0.6279_{\pm 0.0551}$ &$0.0545_{\pm 0.1090}$  &$0.5027_{\pm 0.0409}$  &$0.4184_{\pm 0.0297}$ &$0.6034_{\pm 0.0335}$ &$0.0537_{\pm 0.1074}$\\
            BMVC       &$0.7360_{\pm 0.0095}$  &$0.6302_{\pm 0.0121}$ &$0.7968_{\pm 0.0076}$ &$0.6192_{\pm 0.0154}$  &$0.6038_{\pm 0.0126}$  &$0.5078_{\pm 0.0050}$ &$0.6900_{\pm 0.0118}$ &$0.4530_{\pm 0.0125}$\\
            JSMC       &$0.2463_{\pm 0.0000}$  &$0.0058_{\pm 0.0000}$ &$0.2831_{\pm 0.0000}$ &$0.0004_{\pm 0.0000}$  &$0.1906_{\pm 0.0000}$  &$0.0192_{\pm 0.0000}$ &$0.2414_{\pm 0.0000}$ &$0.0064_{\pm 0.0000}$\\
            UDBGL      &$0.3585_{\pm 0.0000}$  &$0.0178_{\pm 0.0000}$ &$0.3771_{\pm 0.0000}$ &$0.0008_{\pm 0.0000}$  &$0.5421_{\pm 0.0000}$  &$0.5331_{\pm 0.0000}$ &$0.6267_{\pm 0.0000}$ &$0.4113_{\pm 0.0000}$\\
            CL-LSR        &$\textbf{0.9706}_{\pm 0.0000}$  &$\textbf{0.9032}_{\pm 0.0000}$ &$\textbf{0.9705}_{\pm 0.0000}$ &$\textbf{0.9205}_{\pm 0.0000}$  &$0.6361_{\pm 0.0024}$  &$\underline{0.5945}_{\pm 0.0124}$ &$\underline{0.7095}_{\pm 0.0021}$ &$0.4682_{\pm 0.0115}$\\
            \hline
            method &\multicolumn{4}{c}{WebKB} &\multicolumn{4}{c}{wikipedia} \\
            &ACC &NMI &Fs &ARI  &ACC &NMI &Fs &ARI\\
            \cmidrule{2-9}
            MLRR       &$\underline{0.8611}_{\pm 0.0000}$  &$\underline{0.3020}_{\pm 0.0000}$ &$\underline{0.8625}_{\pm 0.0000}$ &$\underline{0.4735}_{\pm 0.0000}$  &$0.4271_{\pm 0.0000}$  &$0.4102_{\pm 0.0003}$ &$0.4902_{\pm 0.0001}$ &$0.2863_{\pm 0.0001}$\\
            DCEL       &$0.5175_{\pm 0.0130}$  &$0.0019_{\pm 0.0022}$ &$0.5592_{\pm 0.0118}$ &$0.0012_{\pm 0.0021}$  &$0.4134_{\pm 0.0298}$  &$0.2825_{\pm 0.0408}$ &$0.4539_{\pm 0.0321}$ &$0.2099_{\pm 0.0394}$\\
            FPMVS-CAG  &$0.6166_{\pm 0.0000}$  &$0.0117_{\pm 0.0000}$ &$0.6014_{\pm 0.0000}$ &$0.0335_{\pm 0.0000}$  &$0.3293_{\pm 0.0041}$  &$0.1726_{\pm 0.0026}$ &$0.2152_{\pm 0.0018}$ &$0.1089_{\pm 0.0012}$\\
            GMC        &$0.7774_{\pm 0.0000}$  &$0.0023_{\pm 0.0000}$ &$0.7605_{\pm 0.0000}$ &$0.0115_{\pm 0.0000}$  &$0.4488_{\pm 0.0000}$  &$0.4170_{\pm 0.0000}$ &$0.4829_{\pm 0.0000}$ &$0.1448_{\pm 0.0000}$\\
            LMVSC      &$0.6839_{\pm 0.0196}$  &$0.0276_{\pm 0.0386}$ &$0.6954_{\pm 0.0153}$ &$0.0144_{\pm 0.0289}$  &$0.5359_{\pm 0.0067}$  &$0.4958_{\pm 0.0057}$ &$\underline{0.6005}_{\pm 0.0051}$ &$0.0697_{\pm 0.1394}$\\
            BMVC       &$0.7147_{\pm 0.0064}$  &$0.1971_{\pm 0.0226}$ &$0.7393_{\pm 0.0059}$ &$0.1833_{\pm 0.0113}$  &$0.1971_{\pm 0.0018}$  &$0.0693_{\pm 0.0024}$ &$0.2121_{\pm 0.0019}$ &$0.0262_{\pm 0.0008}$\\
            JSMC       &$0.5890_{\pm 0.0000}$  &$0.0002_{\pm 0.0000}$ &$0.6266_{\pm 0.0000}$ &$0.0035_{\pm 0.0000}$  &$\textbf{0.5830}_{\pm 0.0000}$  &$\underline{0.5050}_{\pm 0.0000}$ &$\textbf{0.6256}_{\pm 0.0000}$ &$\textbf{0.4284}_{\pm 0.0000}$\\
            UDBGL      &$0.7612_{\pm 0.0000}$  &$0.0924_{\pm 0.0000}$ &$0.7661_{\pm 0.0000}$ &$0.2081_{\pm 0.0000}$  &$0.5108_{\pm 0.0000}$  &$0.4805_{\pm 0.0000}$ &$0.5590_{\pm 0.0000}$ &$0.3580_{\pm 0.0000}$\\
            CL-LSR        &$\textbf{0.8811}_{\pm 0.0000}$  &$\textbf{0.3963}_{\pm 0.0000}$ &$\textbf{0.8852}_{\pm 0.0000}$ &$\textbf{0.5503}_{\pm 0.0000}$  &$\underline{0.5423}_{\pm 0.0006}$  &$\textbf{0.5300}_{\pm 0.0009}$ &$0.5901_{\pm 0.0008}$ &$\underline{0.3895}_{\pm 0.0009}$\\
            \hline
        \end{tabular}
    }
\end{table*}

From Table \ref{tfor4-1-all}, we make the following observations.
\bi
    \item[(i)] The performance of CL-LSR is found to be both satisfactory and highly stable.
    In terms of clustering outcomes, CL-LSR consistently ranks first or second,
    securing the highest number of top positions.
    This consistent superiority serves as empirical evidence affirming the efficacy of the CL-LSR approach.

    \item[(ii)]  Conversely, alternative methodologies exhibit less consistent performance profiles compared to CL-LSR.
        For instance, while GMC achieves optimal results on the Caltech101-7 dataset,
        and JSMC demonstrates enhanced performance on the 3sources and Wikipedia datasets,
        neither of these methods maintains a stable level of excellence across all datasets examined, highlighting their relative inconsistency.

    \item[(iii)] Collectively, these findings illustrate that CL-LSR has successfully harnessed more precise information for executing clustering tasks with greater accuracy.
        The imposition of cardinality constraints on $\{C^v\}_{v=1}^V$ enables CL-LSR to selectively extract reliable information from individual views,
        whereas the low-rank constraint on $C^*$ fosters integration of complementary data across all views. This dual mechanism of constraint enforcement synergistically enhances the clustering process, thereby contributing to the superior performance of our CL-LSR.

\ei

\subsubsection{Visualization and parameter analysis}

\begin{figure}[htbp]
    \centering
    \subfigure[MLRR]{
    \resizebox*{3.6cm}{!}{\includegraphics{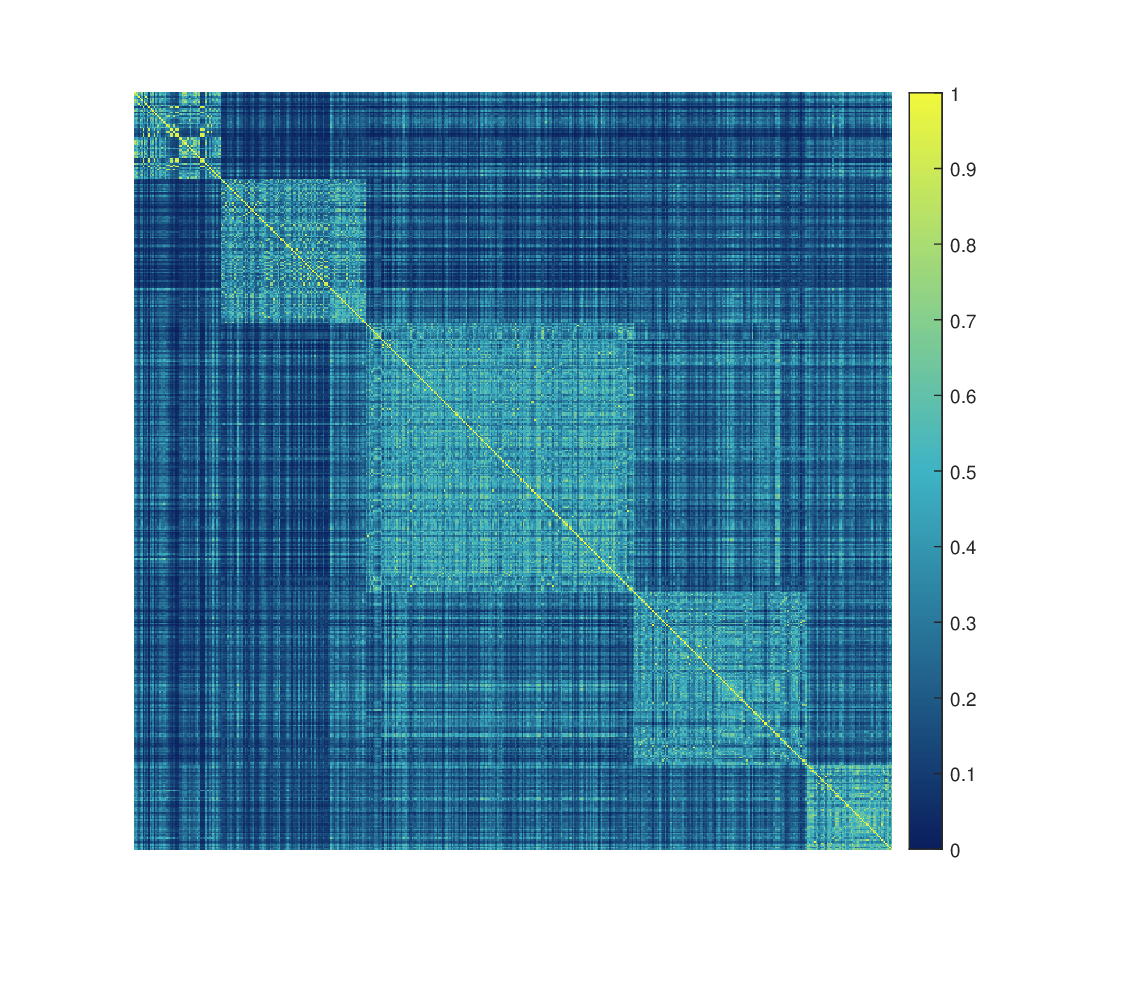}}
    \label{visualization-a}}
    \subfigure[DCEL]{
    \resizebox*{3.6cm}{!}{\includegraphics{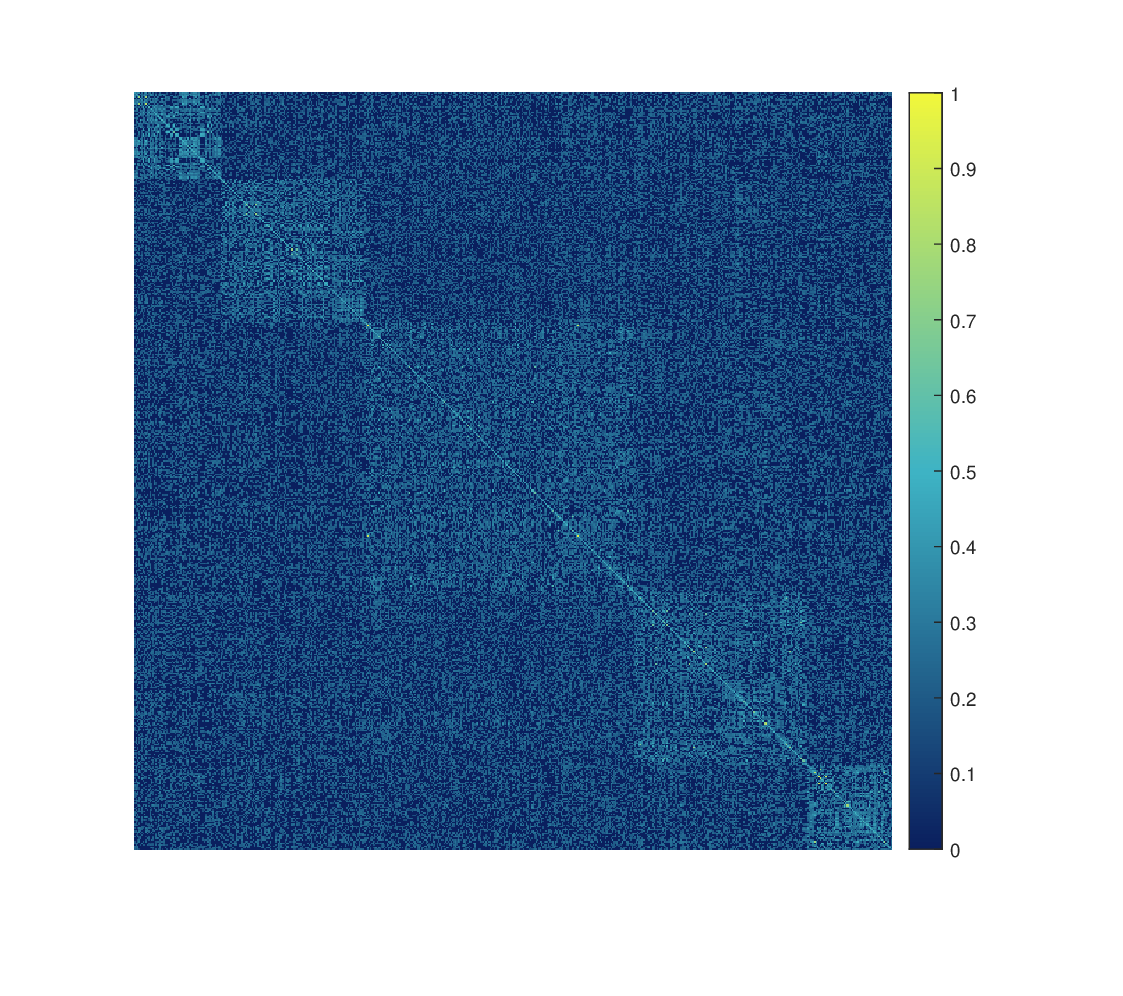}}
    \label{visualization-b}}
    \subfigure[GMC]{
    \resizebox*{3.6cm}{!}{\includegraphics{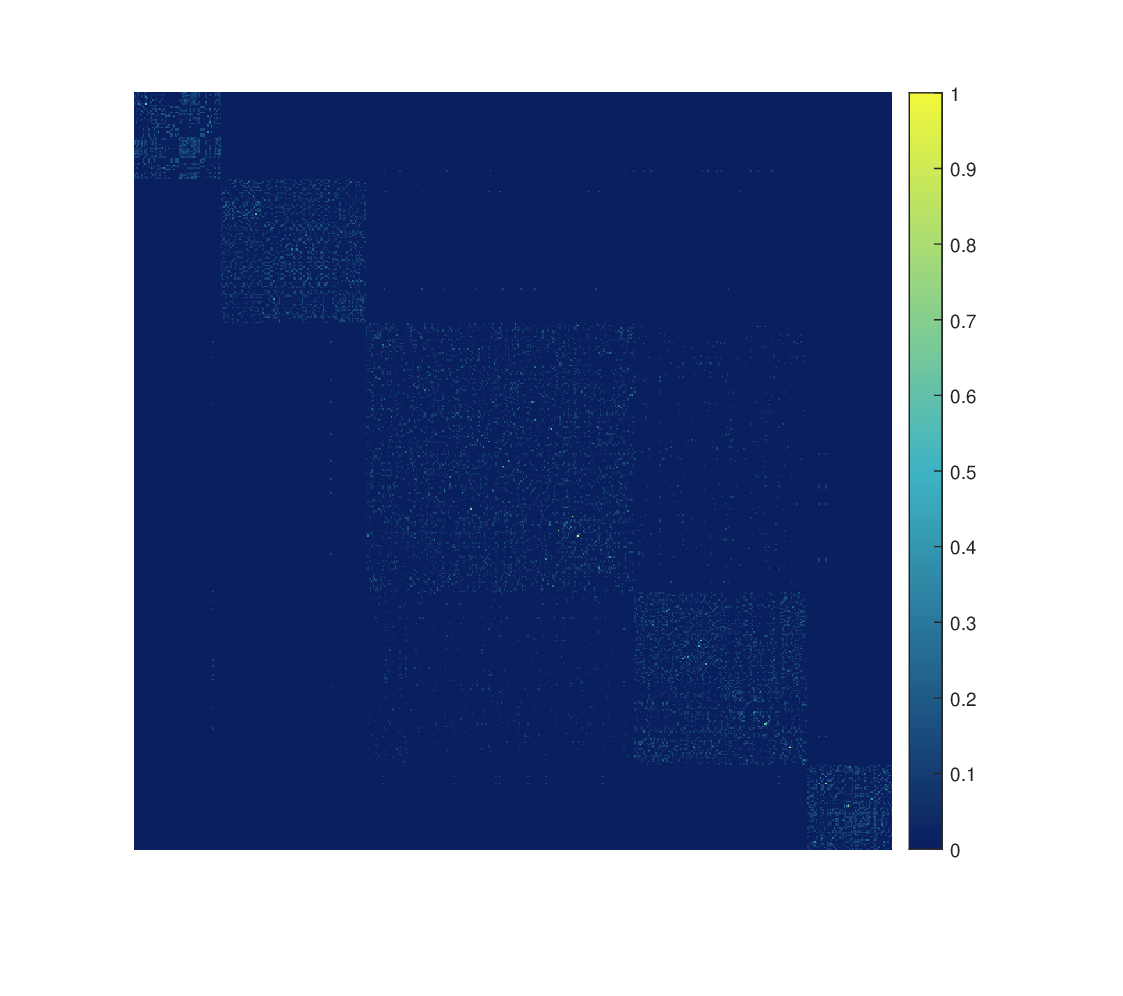}}
    \label{visualization-c}}
    \\
    \subfigure[JSMC]{
    \resizebox*{3.6cm}{!}{\includegraphics{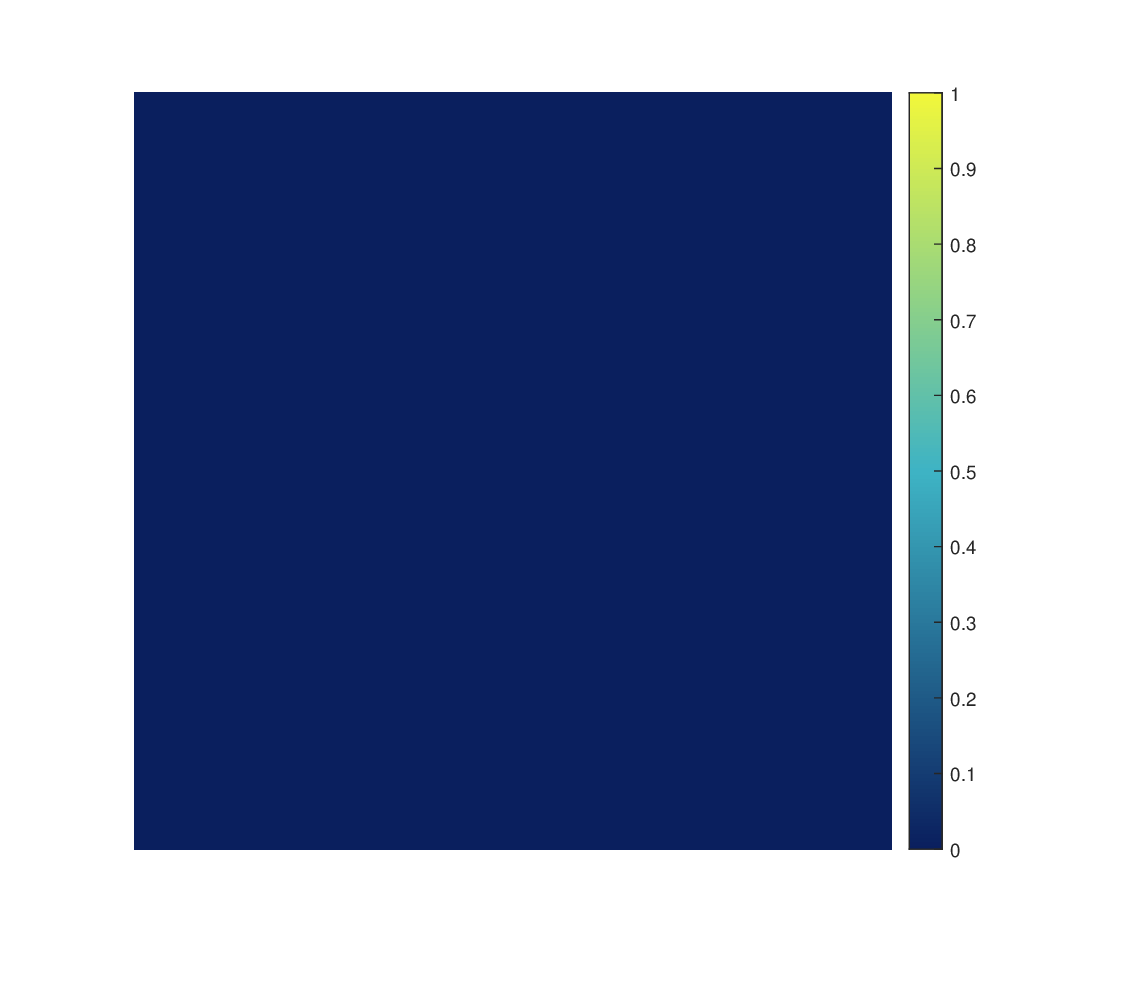}}
    \label{visualization-d}}
    \subfigure[CL-LSR]{
    \resizebox*{3.6cm}{!}{\includegraphics{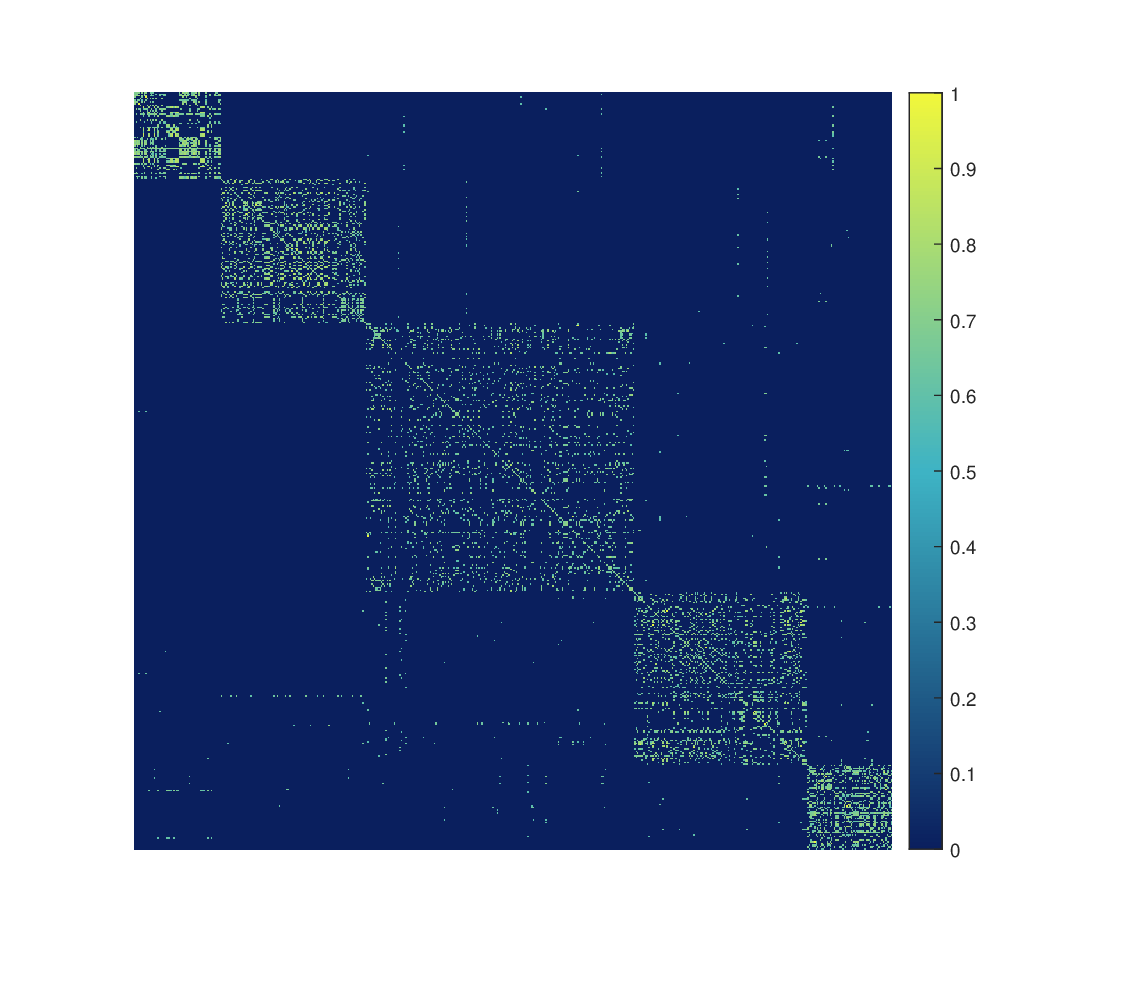}}
    \label{visualization-e}}
    \caption{Visualization of the consensus affinity matrices on BBCsport. \label{visualization}}
\end{figure}

Then, we compare the visualization of the consensus affinity matrix $C^*$
generated by each method.
Specially, we first make a threshold truncation operation on $C^*$,
i.e., the element in $C^*$ is truncated to $0$ if it is smaller than the threshold
$c = 10^{-4}$.
This operation keeps the main similarity information.
After that, we denote the consensus affinity matrix as $\hat C^*$.
Then, for any non-zero elements in $\hat C^*$,
we do the logarithmic transformation as follows
\begin{equation}\label{logc}
    {\hat C^*}_{i,j}=log_{{\hat C^*}_{i,j}}{\max\{\hat C^*_{i,j}\}_{i,j \in [n]}},
\end{equation}
where $\max\{\hat C^*_{i,j}\}_{i,j \in [n]}$ denotes the maximum value of all the elements in $\hat C^*$.
Equation \eqref{logc} maps the elements in $\hat C^*$ to $[0,1]$
and keeps the original size relationship of elements in ${\hat C^*}$ unchanged.
We take BBCsport as an example and repost its numerical results in Fig. \ref{visualization},
which visualizes the consensus affinity matrices.
The light color indicates the high similarity and the dark color represents the low similarity.
As the data in BBCsport are aligned by classes,
the clearer block diagonal structure in the consensus affinity matrix, the better clustering performance.
In addition, one observes that we omit the results of four methods:
LMVSC, BMVC, UDBGL and FPMVS-CAG
since the first three methods do not generate the consensus affinity matrices,
and the code of the last one is encrypted, making it unable to output $C^*$.

From Fig. \ref{visualization}, we make the following observations.
\begin{itemize}
    \item[(i)] As depicted in subfigures \ref{visualization-a}-\ref{visualization-e},
    a majority of methodologies successfully discern the correct data relationships.
    Notably, the JSMC approach deviates from this pattern, lacking a block-diagonal structure that consequently results in inferior performance metrics as documented in Table \ref{tfor4-1-all}.

    \item[(ii)] Turning to MLRR and DCEL, both reveal a block-diagonal structure.
        Nonetheless, they exhibit shortcomings;
        specifically, while the primary diagonal blocks are prominently defined in MLRR and GMC,
        an excessive amount of noise is evident off-diagonal,
        as visually illustrated in subfigures \ref{visualization-a} and \ref{visualization-b}.

    \item[(iii)] In the case of GMC and CL-LSR, their block structures appear less susceptible to noise interference.
        Comparatively, CL-LSR demonstrates more distinctly defined blocks along the main diagonal than GMC,
        a distinction supported by the visual evidence in \ref{visualization-c} and \ref{visualization-e}. This visual assessment corroborates the enhanced efficacy of the CL-LSR method.
\end{itemize}

\begin{figure}[htpb]
    \centering
    \subfigure[ACC]{
    \resizebox*{3.8cm}{!}{\includegraphics[scale=0.2]{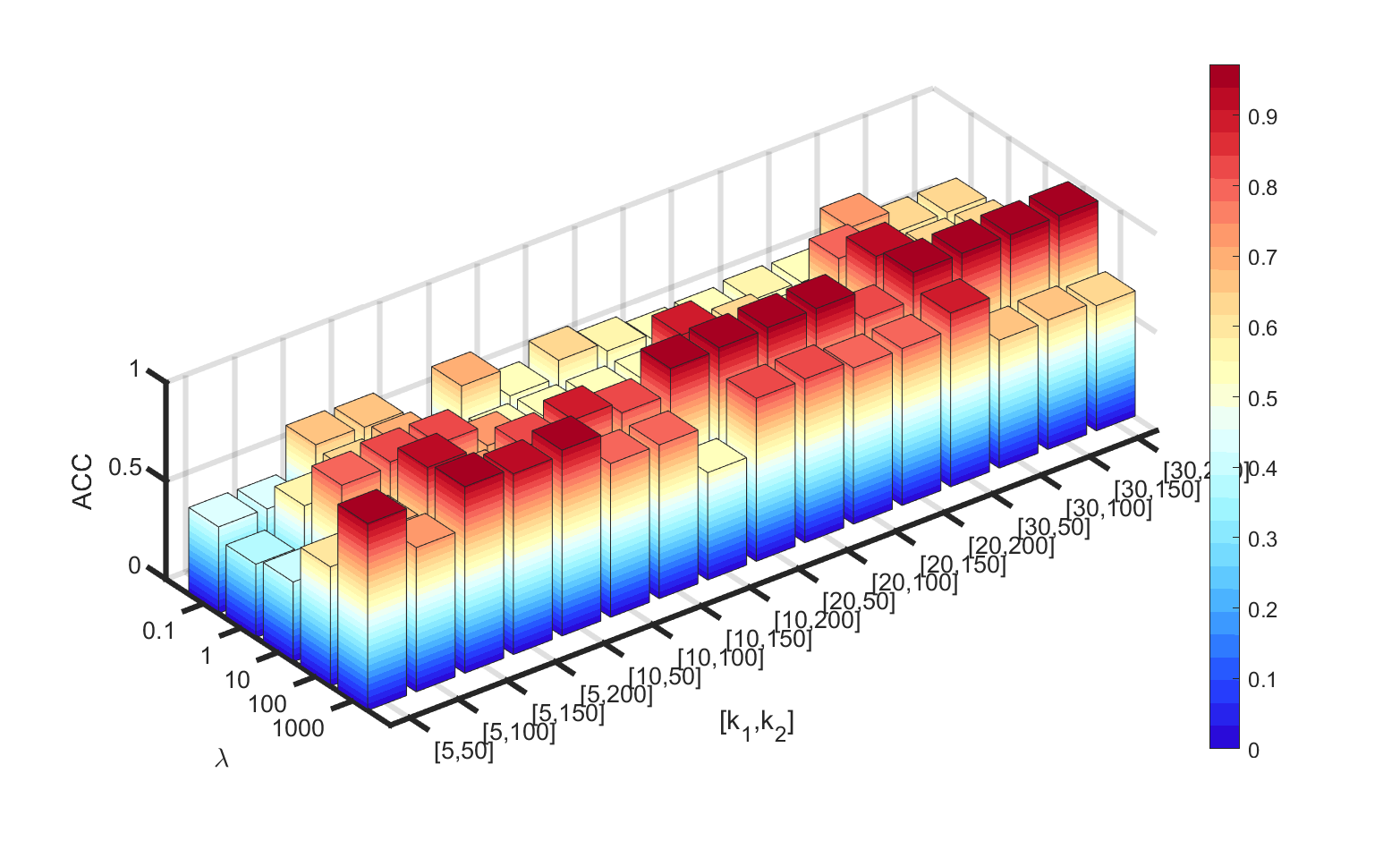}}
    \label{parabbcsport-a}}
    \subfigure[NMI]{
    \resizebox*{3.8cm}{!}{\includegraphics[scale=0.2]{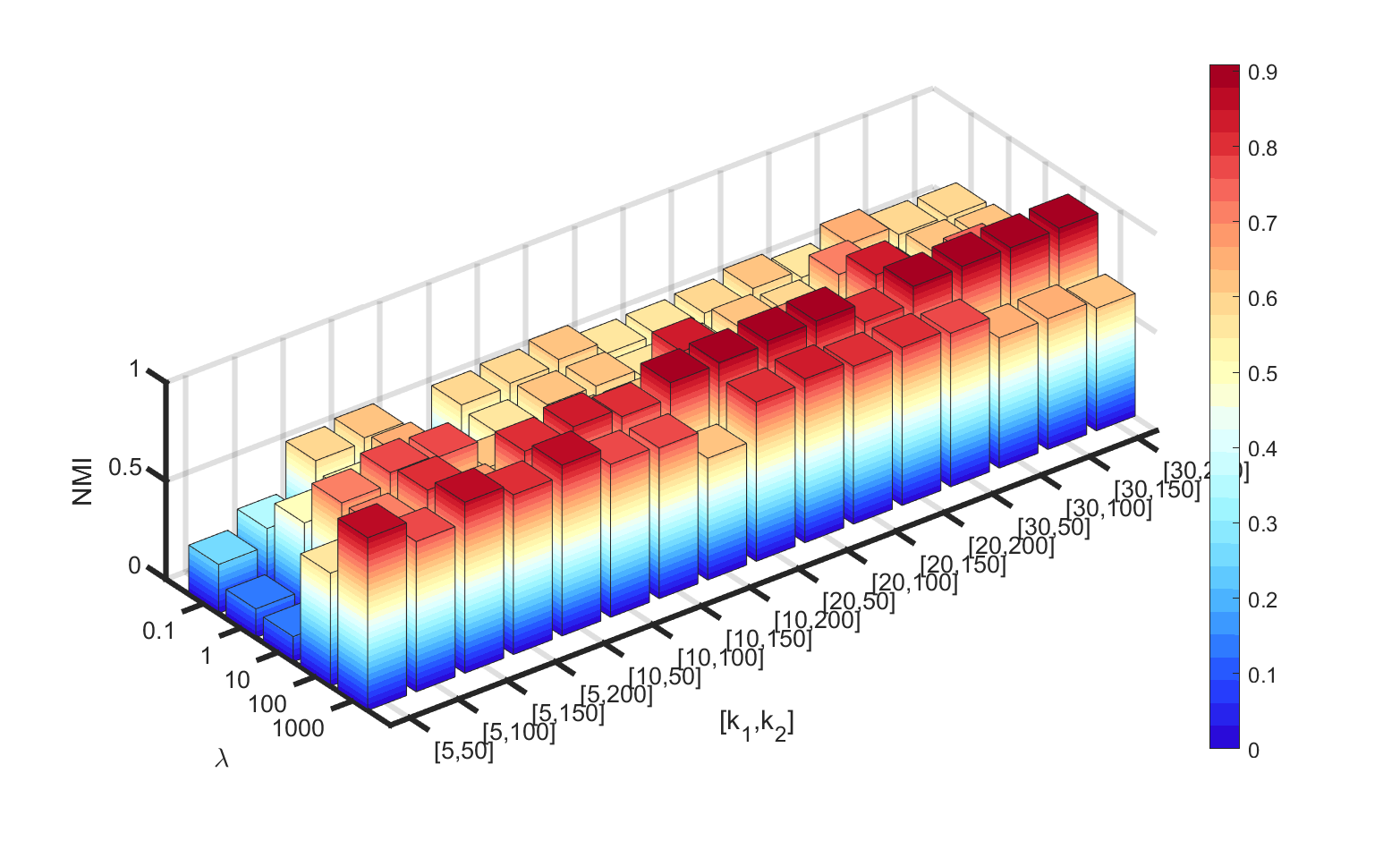}}
    \label{parabbcsport-b}} \\
    \subfigure[Fs]{
    \resizebox*{3.8cm}{!}{\includegraphics[scale=0.2]{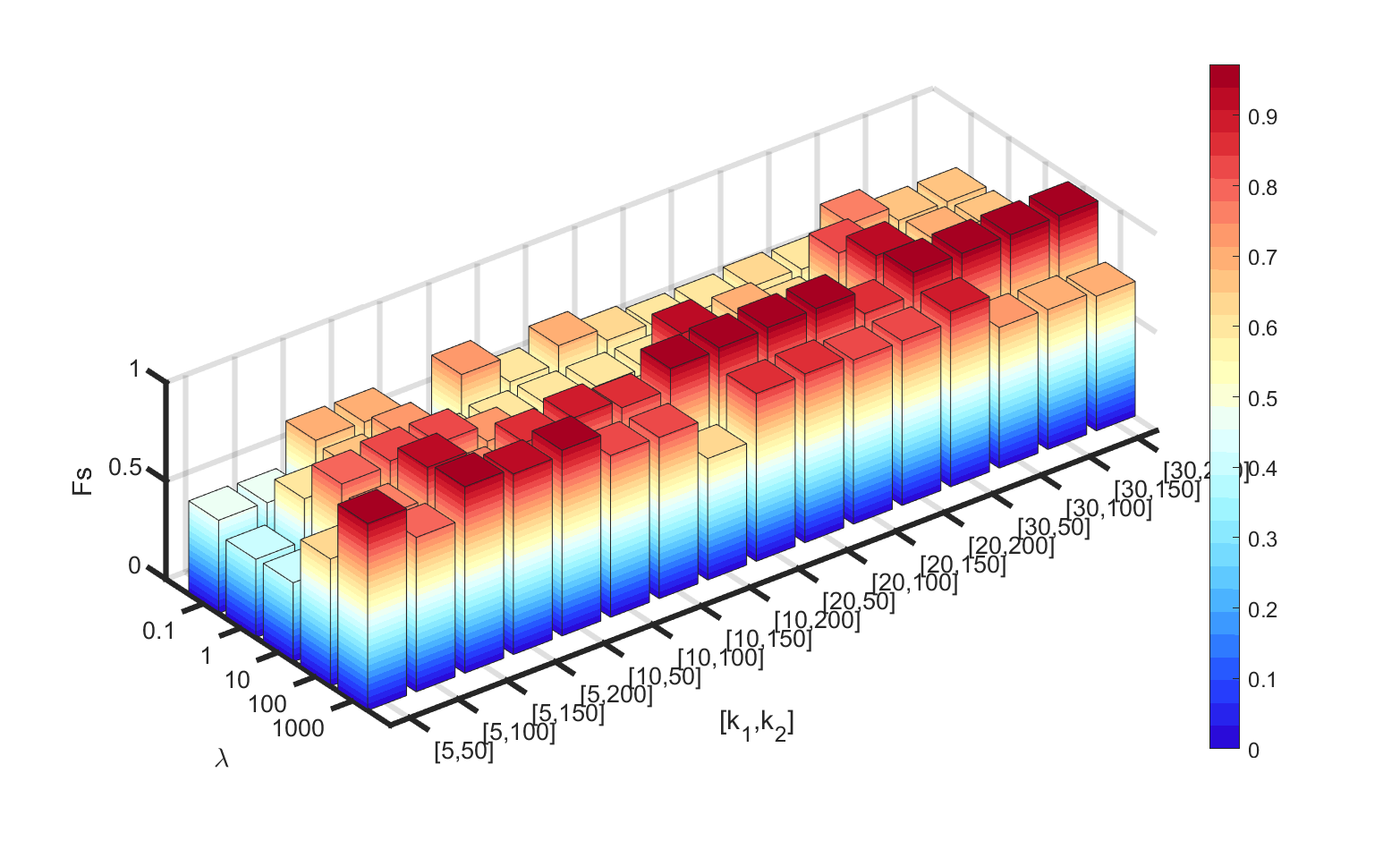}}
    \label{parabbcsport-c}}
    \subfigure[ARI]{
    \resizebox*{3.8cm}{!}{\includegraphics[scale=0.2]{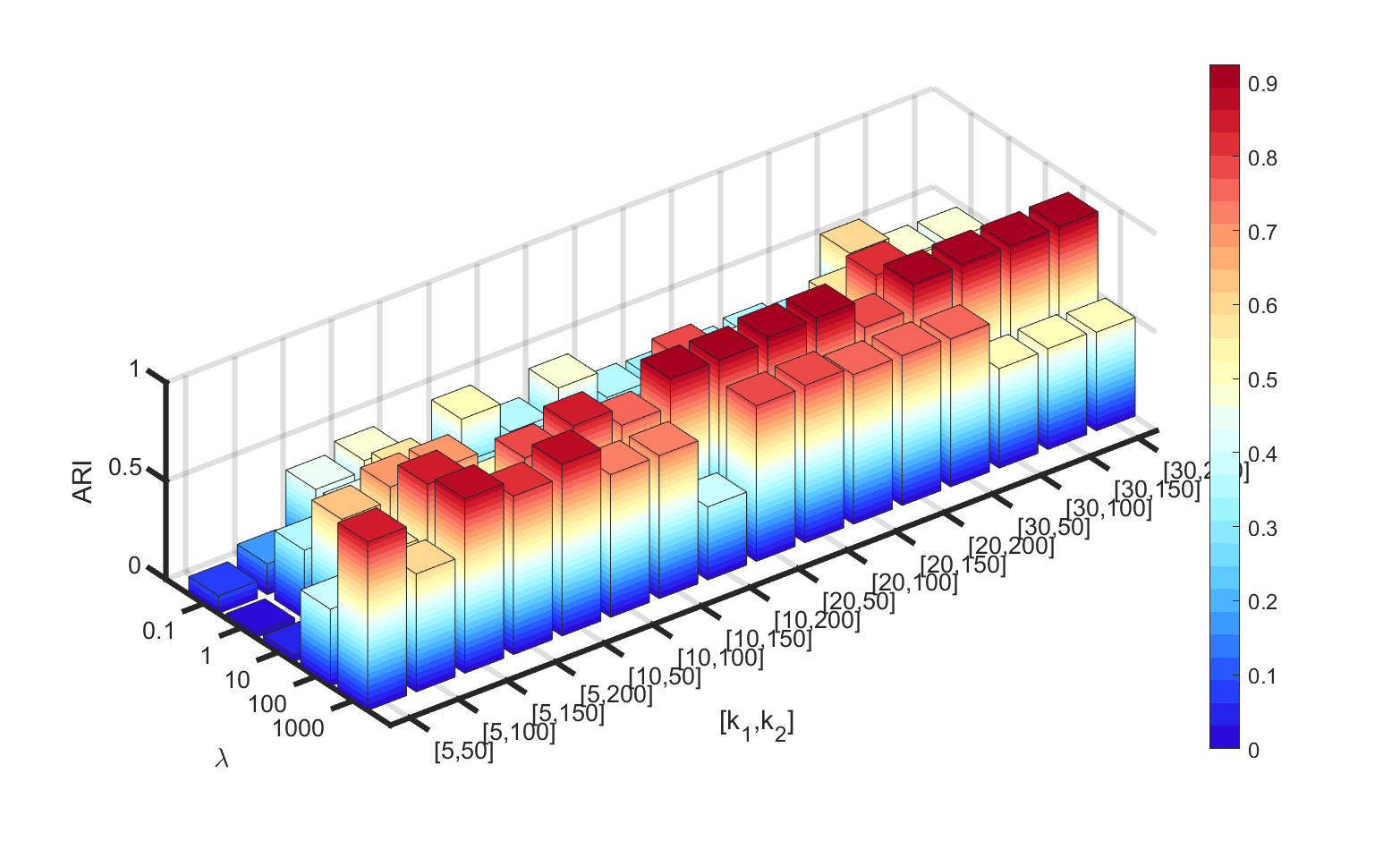}}
    \label{parabbcsport-d}}
    \caption{Parameter effects on BBCsport}
    \label{parabbcsport}
\end{figure}

Next, we illustrate the impact of parameters on the performance of CL-LSR on BBCsport dataset.
Specially, we report the heat maps of the aforementioned four clustering criteria
in Fig. \ref{parabbcsport} under various combinations of parameters $(\lambda, k_1, k_2)$.
The changing sets of $\lambda, k_1$ and $k_2$ are $\{0.1, 1, 10, 100, 1000\}$,
$\{5, 10, 20, 30\}$ and $\{10 k_c, 20 k_c, 30 k_c, 40 k_c\}$, respectively,
where $k_c$ is the number of the predefined clusters.
For no further elaboration,
the other figures of parameter analysis on the remaining datasets are reported in Appendix \ref{secA1}.
Notice that, not all combinations of $(\lambda, k_1, k_2)$ match the datasets.
We can only test all the allowed combinations.
For example, there are no allowed parameter combinations of $(k_1, k_2)$ for ORL.
Thus, we set $k_1=5$ and $k_2 \in \{100, 200\}$ on ORL.

From Fig. \ref{parabbcsport} and Figs. \ref{paraorl}-\ref{parawebkb} in Appendix \ref{secA1},
we make the following observations.

\bi

\item[(i)] The performance of CL-LSR demonstrates a notable insensitivity to variations in parameter settings across the ORL, 3sources, Caltech101-7, and Wikipedia datasets.
    In these four datasets, it is observed that extensive exploration of parameter space,
    specifically broad combinations of $(\lambda, k_1, k_2)$, tends to enhance clustering outcomes, suggesting a robustness of CL-LSR to parameter fluctuations within this context.
   

\item[(ii)] Conversely, distinct patterns emerge when examining the BBCsport and WebKB datasets.
Here, the impact of parameter adjustments varies significantly.
For BBCsport, the parameter $\lambda$ exerts a substantial influence,
with clustering efficacy markedly improving as $\lambda$ increases,
while the combined effect of $(k_1, k_2)$ is less pronounced.
In the case of WebKB, the interplay between $k_1$ and $k_2$ becomes crucial,
where larger pairings of $(k_1, k_2)$ generally correlate with superior clustering results.
However, higher values of $\lambda$ tend to degrade performance,
likely due to the simplistic two-cluster structure of WebKB.
Unlike the intricate structures of view-specific affinity matrices, the global structure demands greater attention.
This underscores the dataset-specific nuances in parameter sensitivity.

\item[(iii)] Collectively, these findings illustrate that a universal parameter configuration optimal for all datasets does not exist.
    Nonetheless, a prevailing trend suggests that, in the majority of scenarios, augmenting the values of parameters $k_1$ and $k_2$ proves advantageous for clustering performance.
    This tendency can be attributed to the enrichment of structural information encapsulated by CL-LSR at higher $(k_1, k_2)$ values,
which in turn fosters an enhanced clustering effect.
In addition, for most datasets, under higher values of $(k_1, k_2)$,
a relatively large $\lambda$ tends to enhance performance.
However, for simpler datasets like WebKB, a smaller $\lambda$ is recommended.
These insights underscore the importance of tailored parameter tuning in harnessing the full potential of multiview clustering methodologies like CL-LSR.
\ei

\subsection{Convergence curve}

\begin{figure}[!htb]
    \centering
    \subfigure[POF on ORL]{
    \resizebox*{2.8cm}{!}{\includegraphics[scale=0.3]{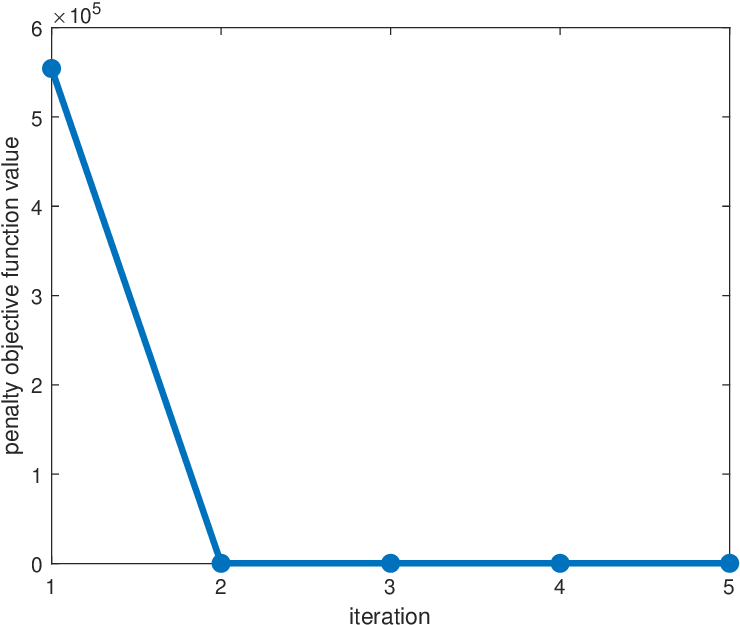}}
    \label{convergence-a}}
    \subfigure[OEF on ORL]{
    \resizebox*{2.8cm}{!}{\includegraphics[scale=0.3]{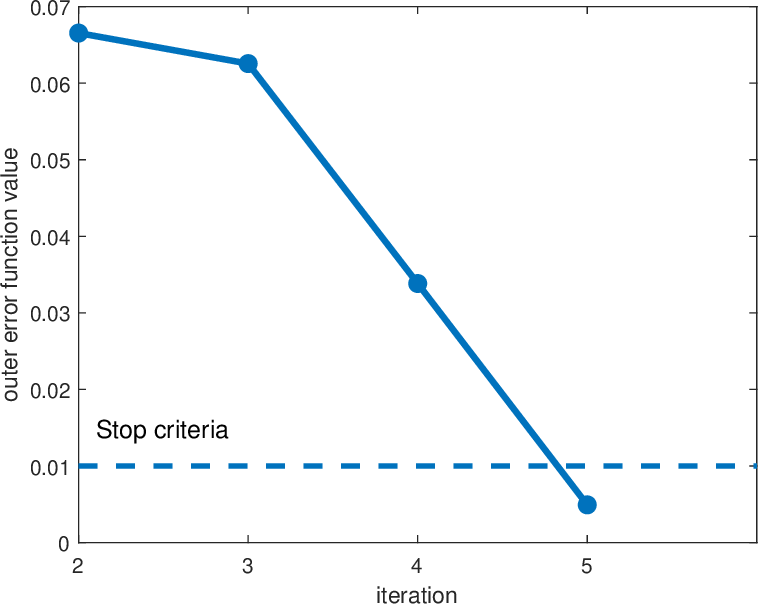}}
    \label{stop-a}}
    \subfigure[POF on 3sources]{
    \resizebox*{2.8cm}{!}{\includegraphics[scale=0.3]{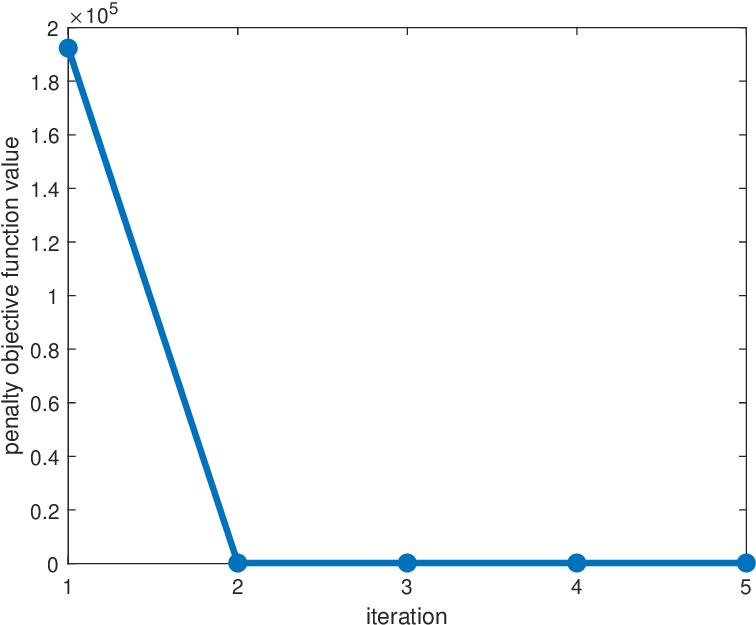}}
    \label{convergence-b}}
    \subfigure[OEF on 3sources]{
    \resizebox*{2.8cm}{!}{\includegraphics[scale=0.3]{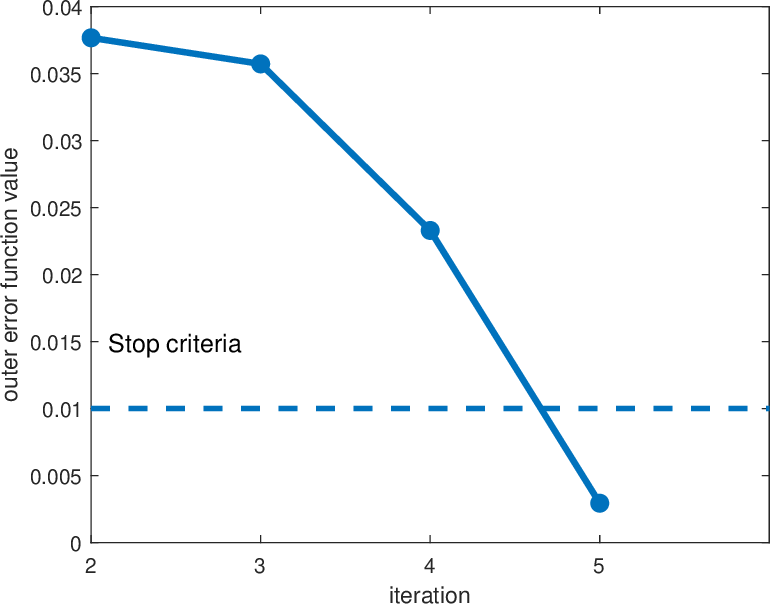}}
    \label{stop-b}} \\

    \subfigure[POF on BBCsport]{
    \resizebox*{2.8cm}{!}{\includegraphics[scale=0.3]{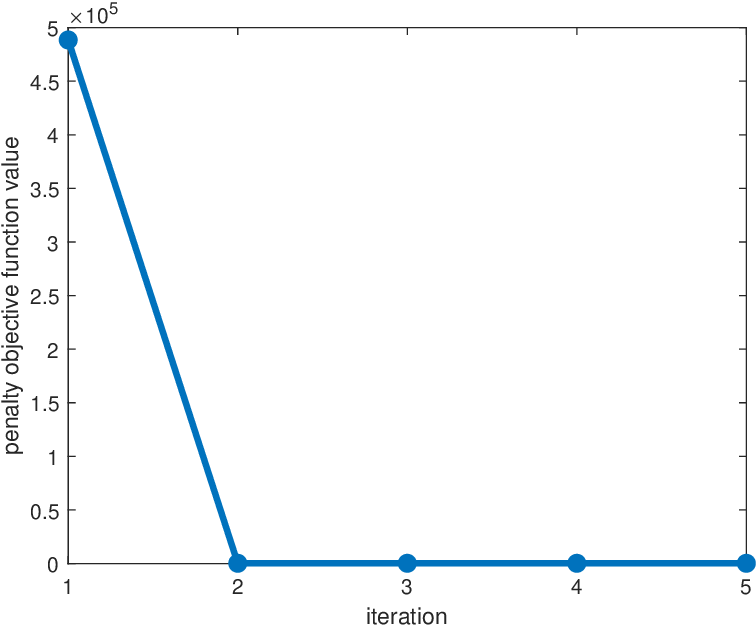}}
    \label{convergence-c}}
    \subfigure[OEF on BBCsport]{
    \resizebox*{2.8cm}{!}{\includegraphics[scale=0.3]{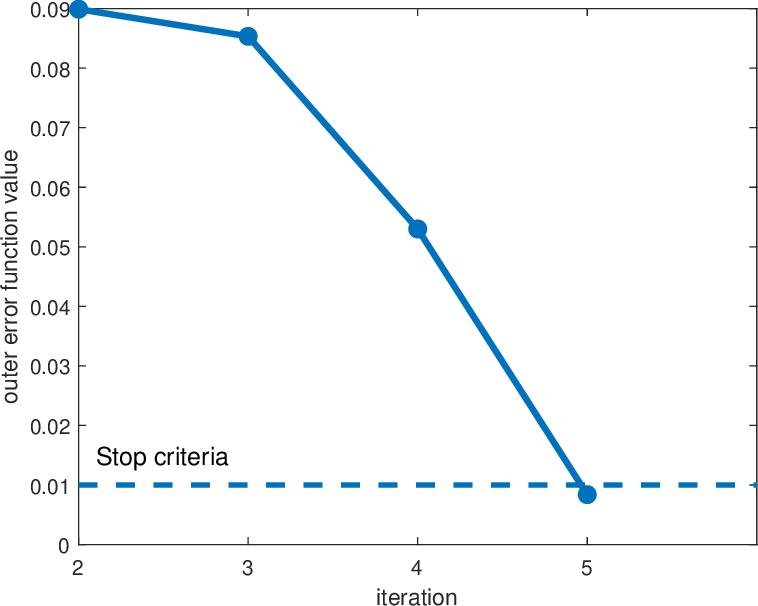}}
    \label{stop-c}}
    \subfigure[POF on Caltech101-7]{
    \resizebox*{2.9cm}{!}{\includegraphics[scale=0.3]{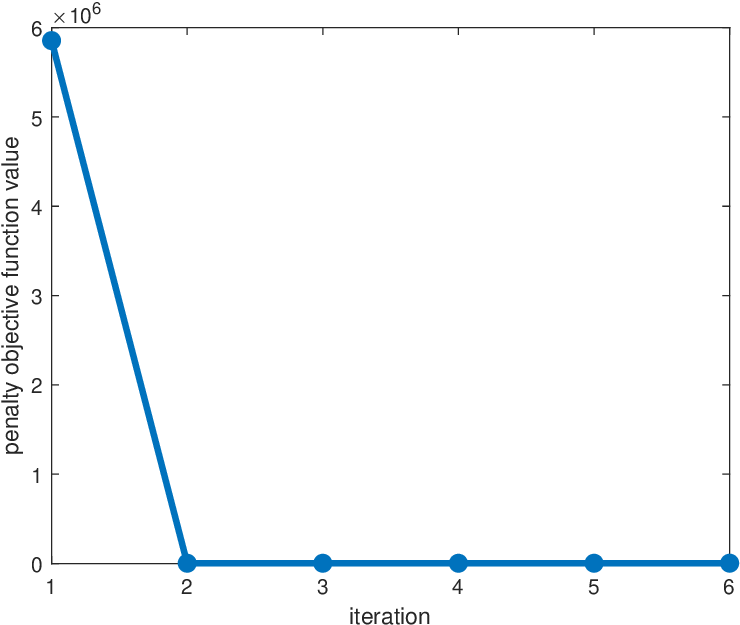}}
    \label{convergence-d}}
    \subfigure[OEF on Caltech101-7]{
    \resizebox*{2.9cm}{!}{\includegraphics[scale=0.3]{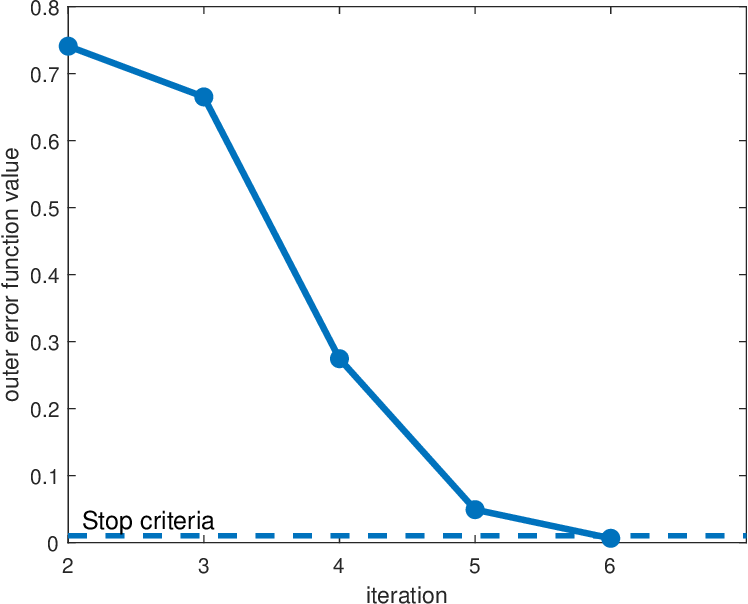}}
    \label{stop-d}} \\

    \subfigure[POF on WebKB]{
    \resizebox*{2.8cm}{!}{\includegraphics[scale=0.3]{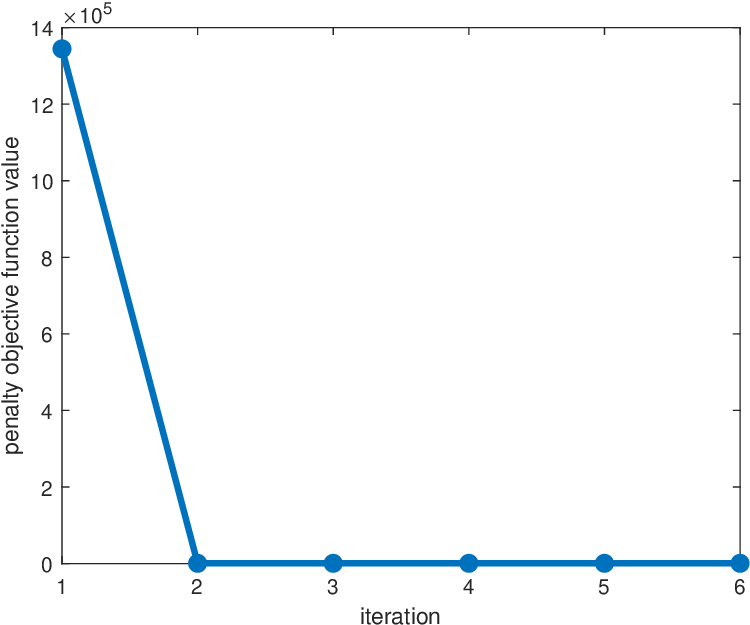}}
    \label{convergence-e}}
    \subfigure[OEF on WebKB]{
    \resizebox*{2.8cm}{!}{\includegraphics[scale=0.3]{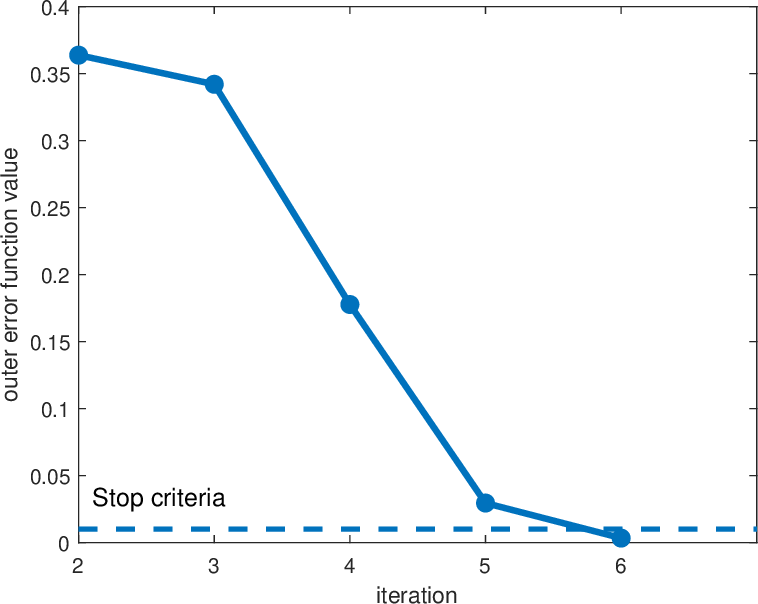}}
    \label{stop-e}}
    \subfigure[POF on wikipedia]{
    \resizebox*{2.8cm}{!}{\includegraphics[scale=0.3]{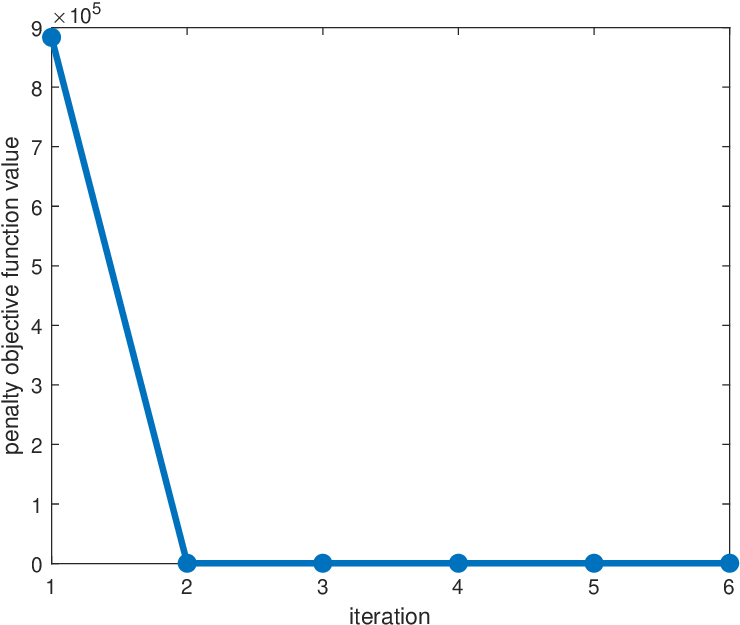}}
    \label{convergence-f}}
    \subfigure[OEF on wikipedia]{
    \resizebox*{2.8cm}{!}{\includegraphics[scale=0.3]{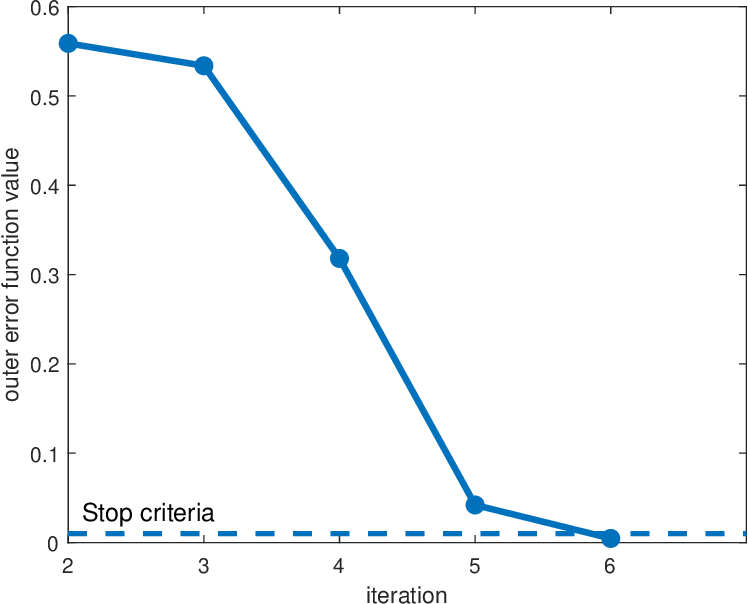}}
    \label{stop-f}}\\

    \caption{Convergence curves of the objective function and outer error function on the six datasets.}
    \label{convergence}
\end{figure}

Finally, we report in Fig. \ref{convergence} the curves of change in
the penalty objective function (POF) and the outer error function (OEF), i.e., $q_\sigma(\{C^v\}_{v=1}^V, C^*)$ and $\max\{\|C^v_k-C^*_k\|_F\}_{v=1}^V$,
with iterations on the six datasets.
Specially, the first and the third columns in Fig. \ref{convergence} show the trend of the penalty objective function values with outer iterations,
while the second and the fourth ones show the corresponding changes of the outer error function values from the second outer iteration to convergence.

From the first and the third columns in Fig. \ref{convergence},
one observes that the values of the penalty objective function decrease sharply after the first two outer iterations
thanks to the closed-form solutions of two subproblems.
Then, as the penalty parameter $\sigma$ increases rapidly,
the decrease in the value of the objective function becomes slower.
However, in this stage, our method is still effective.
This is because the values of error function remain rapidly decreasing until the algorithm converges within finite steps as shown in the second and the fourth columns.
Hence, summarizing the above results, our proposed AQP method is remarkably efficient.

\section{Concluding Remarks}

This paper proposed a novel  joint learning model for multiview subspace clustering,
termed cardinality-constrained low-rank least squares regression (CL-LSR).
In the CL-LSR, we utilized the cardinality constraints under each view to obtain reliable structure information, and employed the low-rank constraint to pursue a consensus
affinity matrix with a clear structure when integrating the complementary information.
For the solution of CL-LSR, an efficient alternating quadratic penalty (AQP) method was proposed, where each block had explicit iterations.
Moreover, the global convergence analysis of the AQP method was established.
Finally, we compared the proposed method with eight state-of-the-art clustering methods,
and obtained favorable performance on six datasets.

In addition, we shall mention that the proposed AQP method in this paper can be used
to solve the problems arising in image segmentation, multiple kernel learning, ect.
The corresponding theoretical results still hold.

\bmhead{Acknowledgements}

This work was supported by National Natural Science Foundation of China (Nos. 12301405, 62406261), Shaanxi Fundamental Science Research Project for Mathematics and Physics (No. 23JSZ010) 
and Fundamental Research Funds for Central Universities of China (No. ZYTS25201).










\newpage

\begin{appendices}

\section{Parameters analysis on the rested five datasets}\label{secA1}

Other five figures for the rested five datasets are given in this appendix.
Parameters settings are given as the paper section 4.2.2.
\begin{figure}[H]
    \centering
    \subfigure[ACC]{
    \resizebox*{3.4cm}{!}{\includegraphics{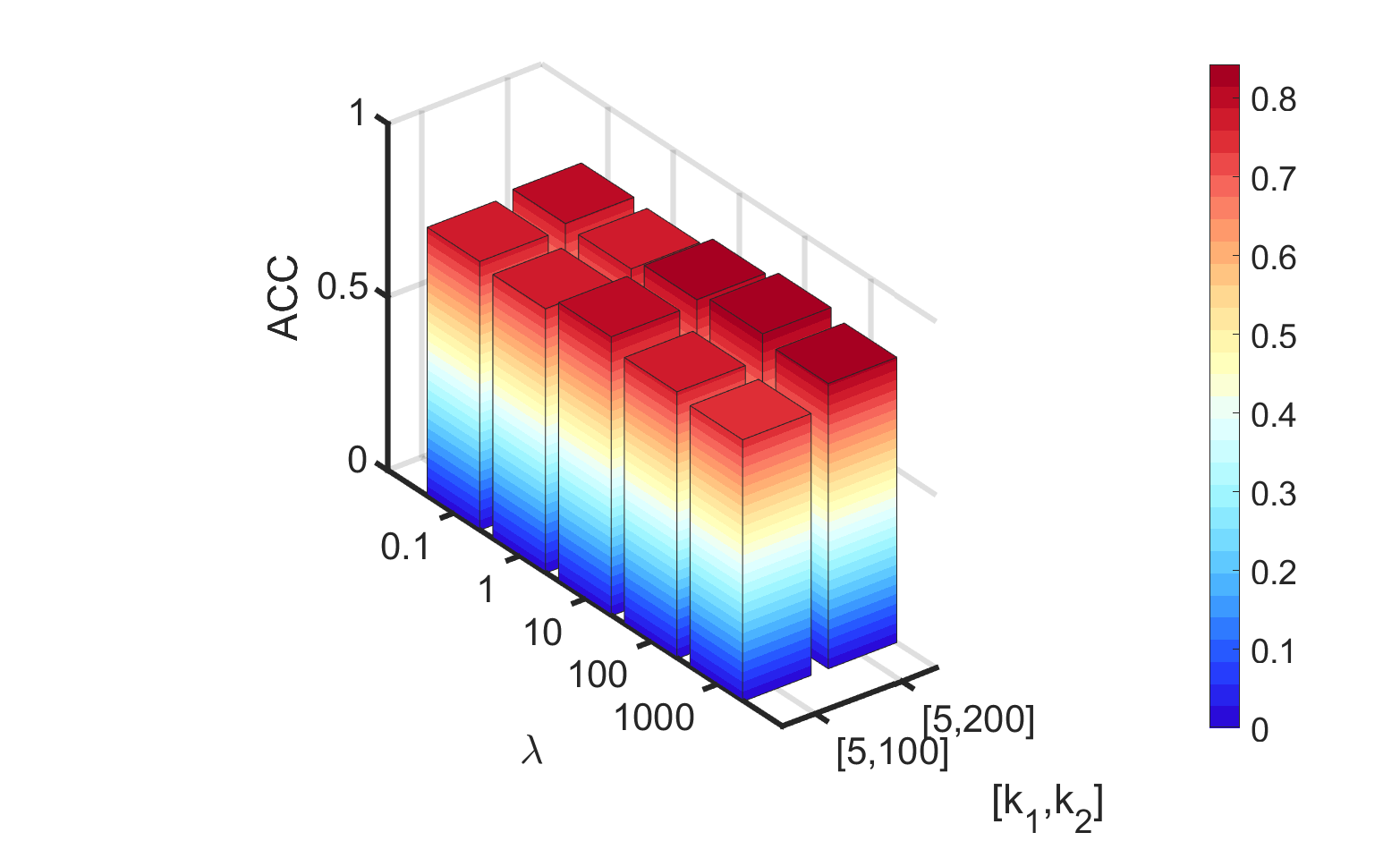}}
    \label{paraorl-a}}
    \subfigure[NMI]{
    \resizebox*{3.4cm}{!}{\includegraphics[scale=0.2]{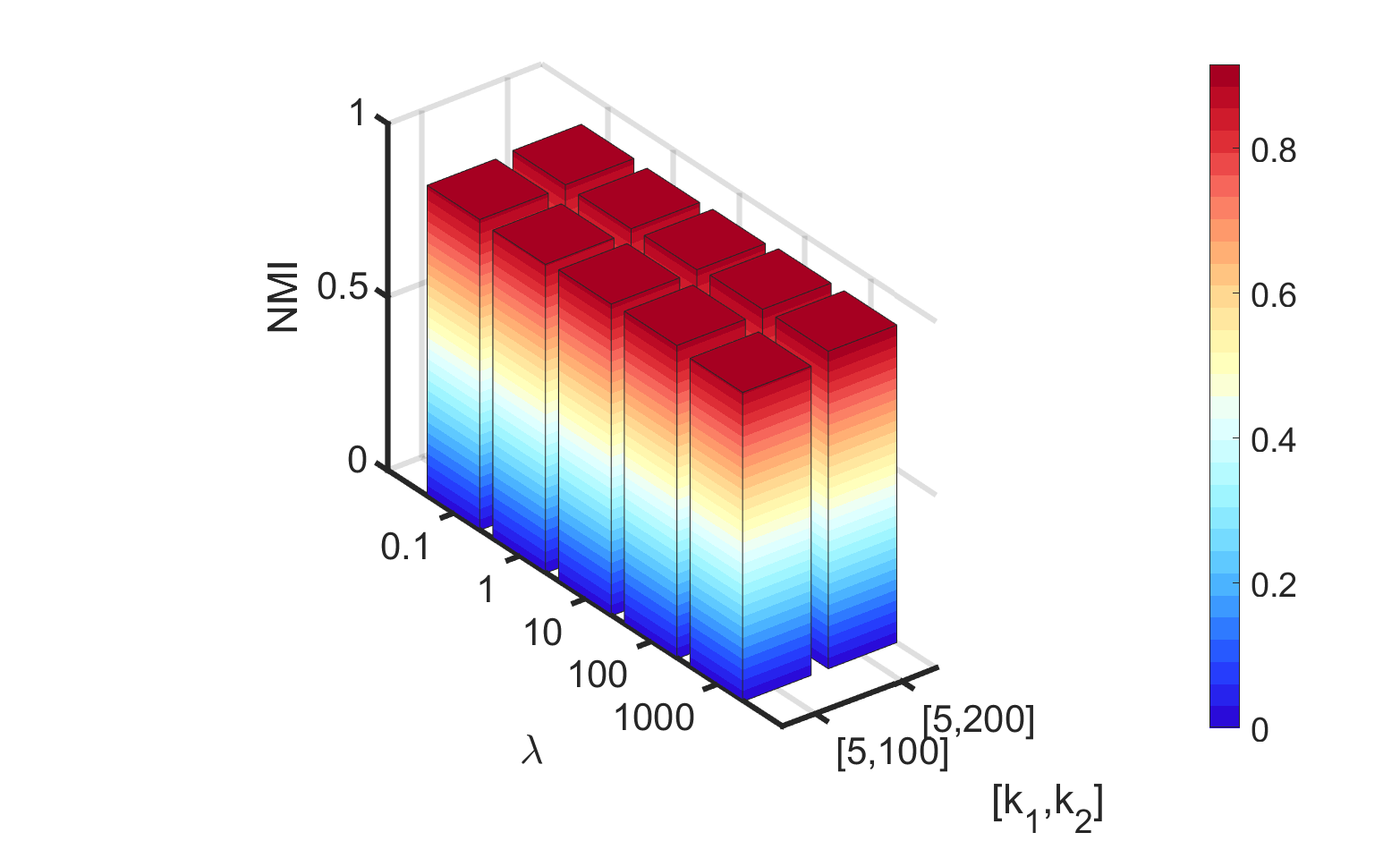}}
    \label{paraorl-b}} \\
    \subfigure[Fs]{
    \resizebox*{3.4cm}{!}{\includegraphics[scale=0.2]{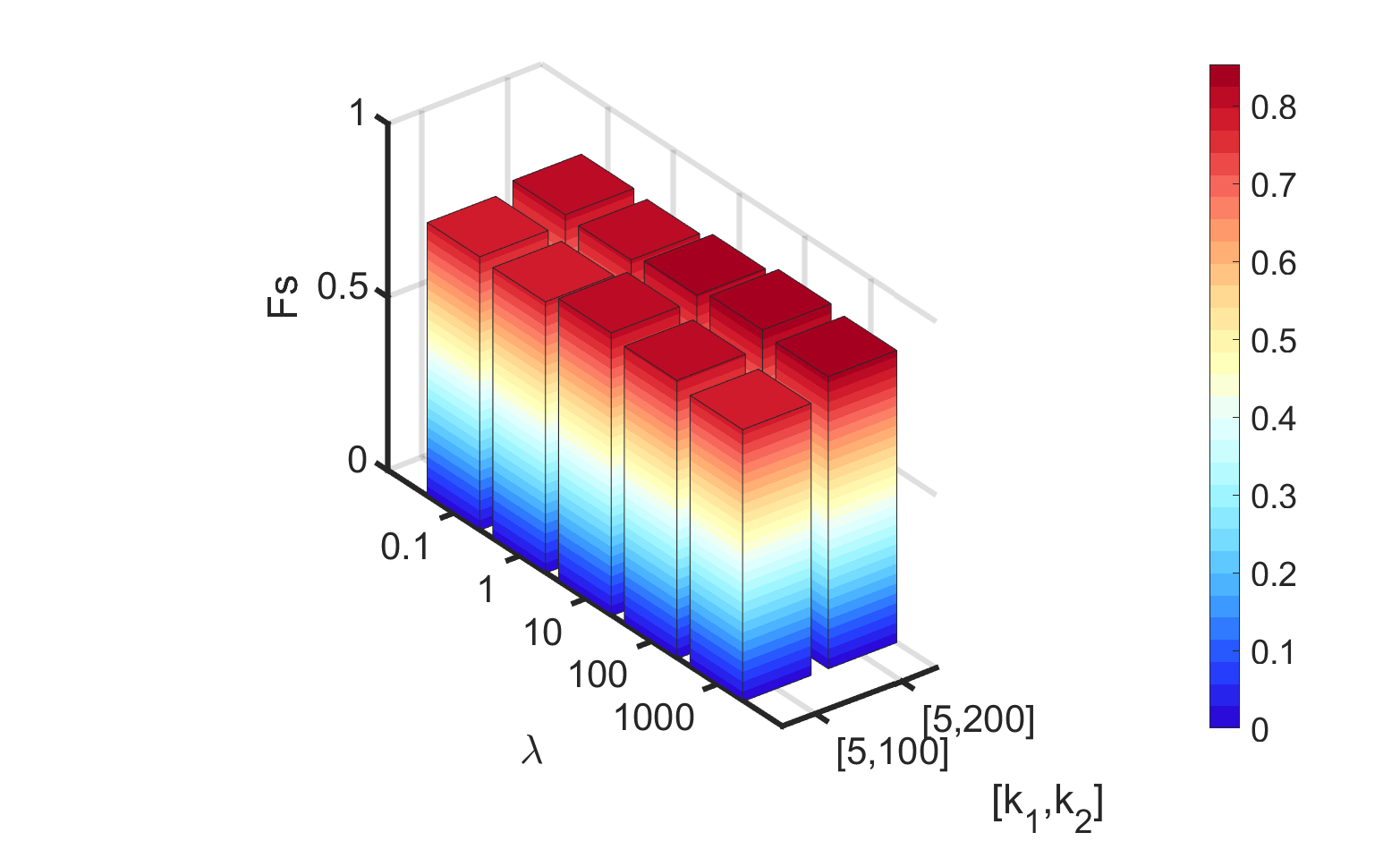}}
    \label{paraorl-c}}
    \subfigure[ARI]{
    \resizebox*{3.4cm}{!}{\includegraphics[scale=0.2]{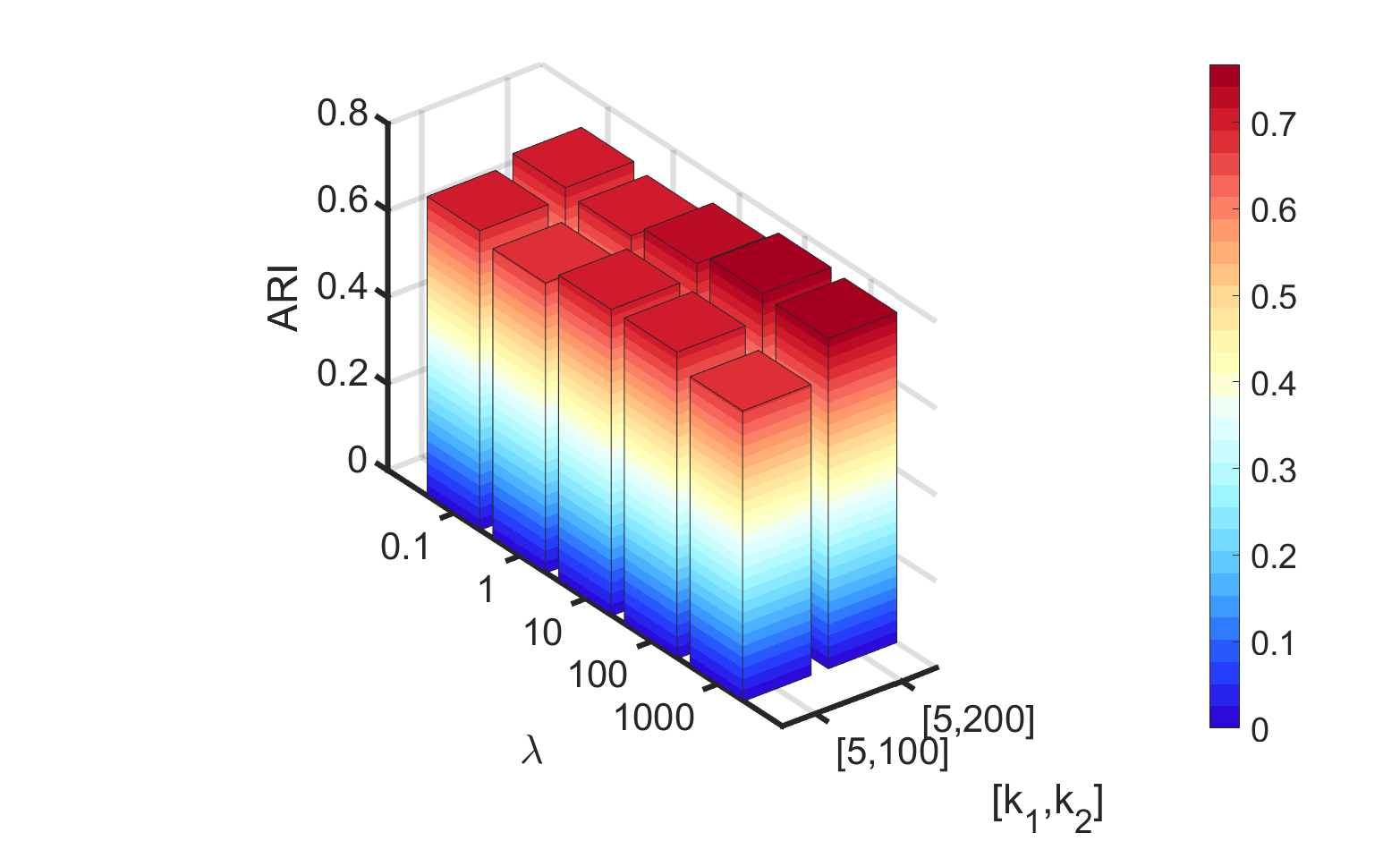}}
    \label{paraorl-d}}
    \caption{Parameter effects on ORL.}
    \label{paraorl}
\end{figure}


\begin{figure}[H]
    \centering
    \subfigure[ACC]{
    \resizebox*{3.4cm}{!}{\includegraphics[scale=0.2]{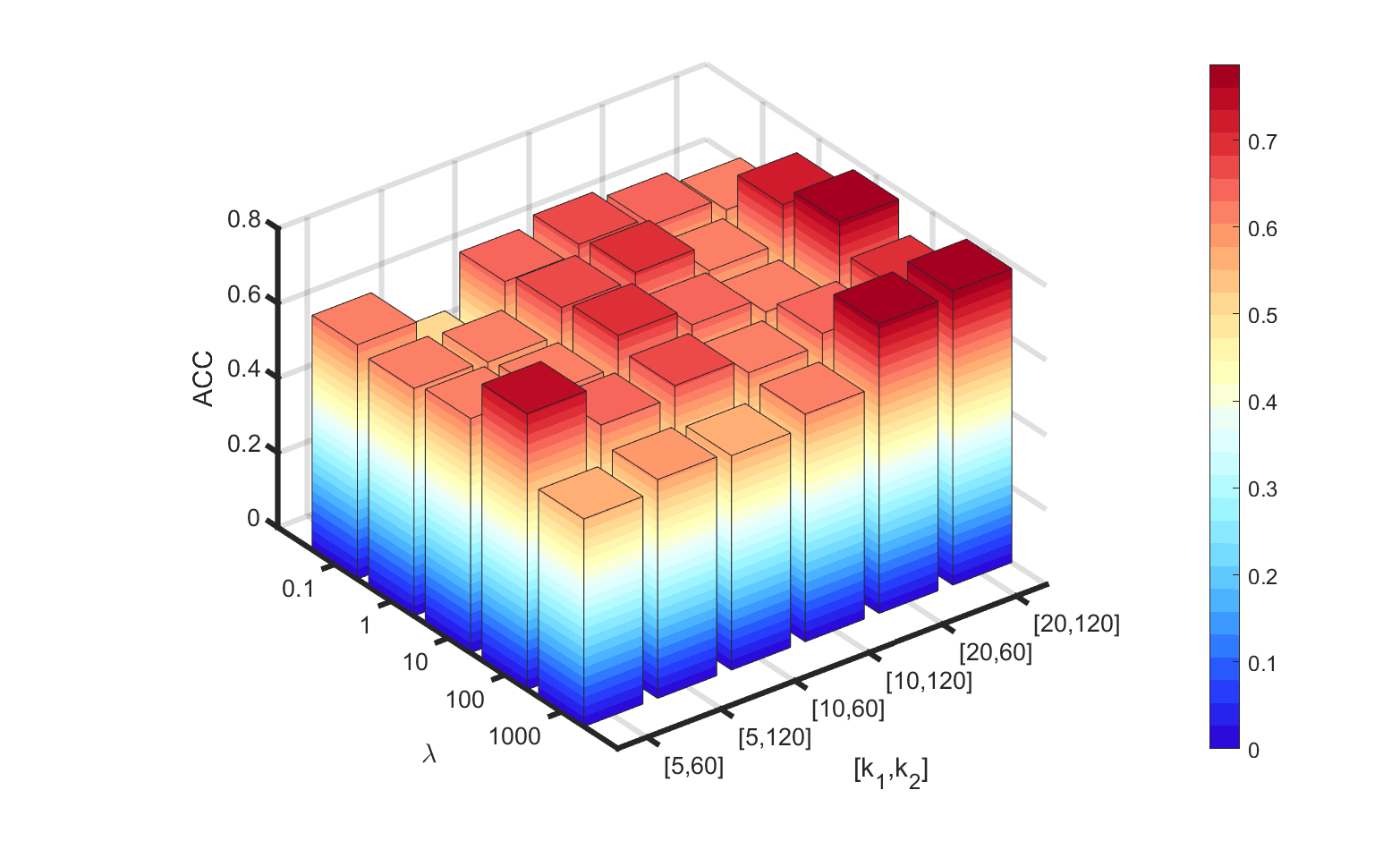}}
    \label{para3sources-a}}
    \subfigure[NMI]{
    \resizebox*{3.4cm}{!}{\includegraphics[scale=0.2]{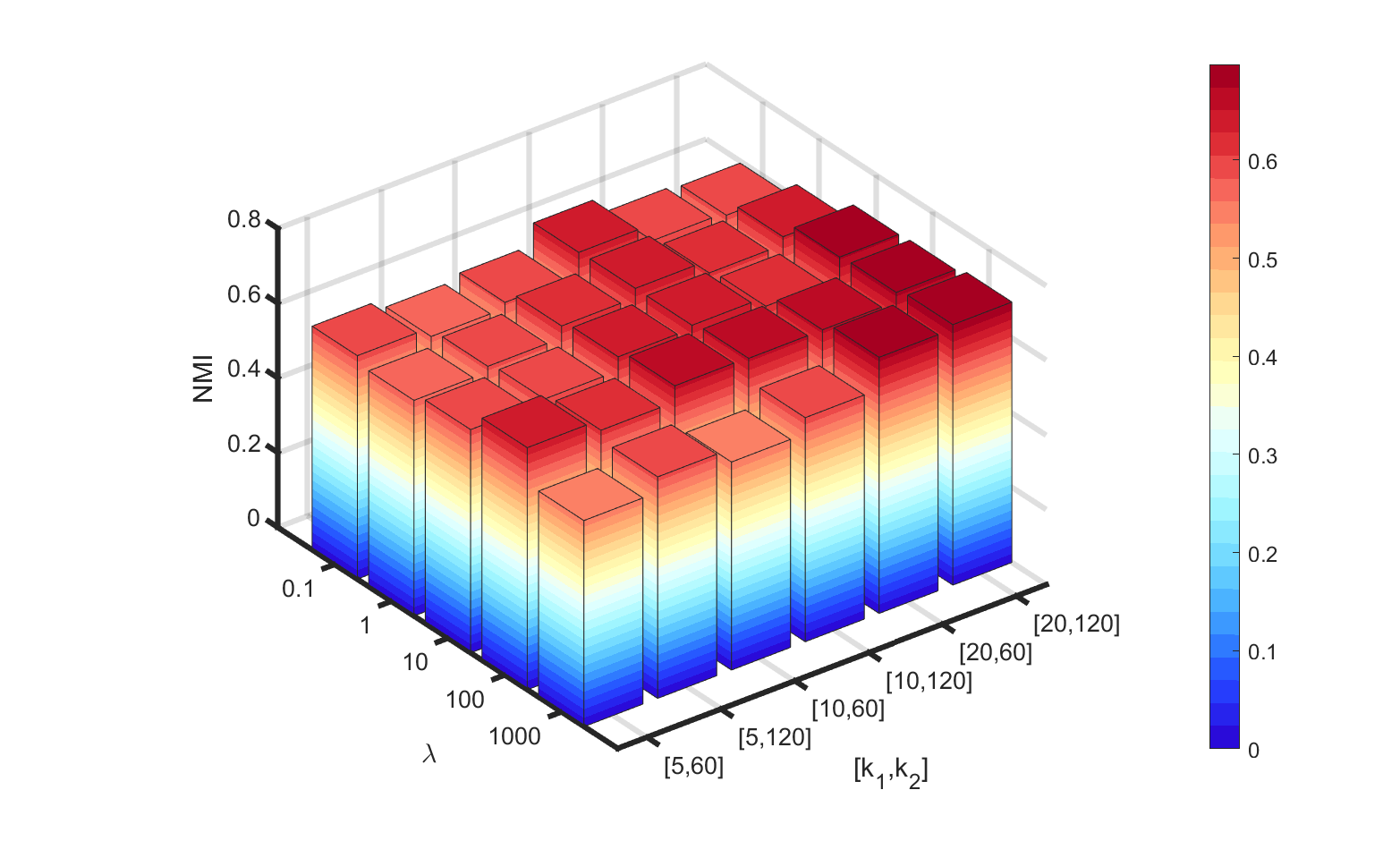}}
    \label{para3sources-b}} \\
    \subfigure[Fs]{
    \resizebox*{3.4cm}{!}{\includegraphics[scale=0.2]{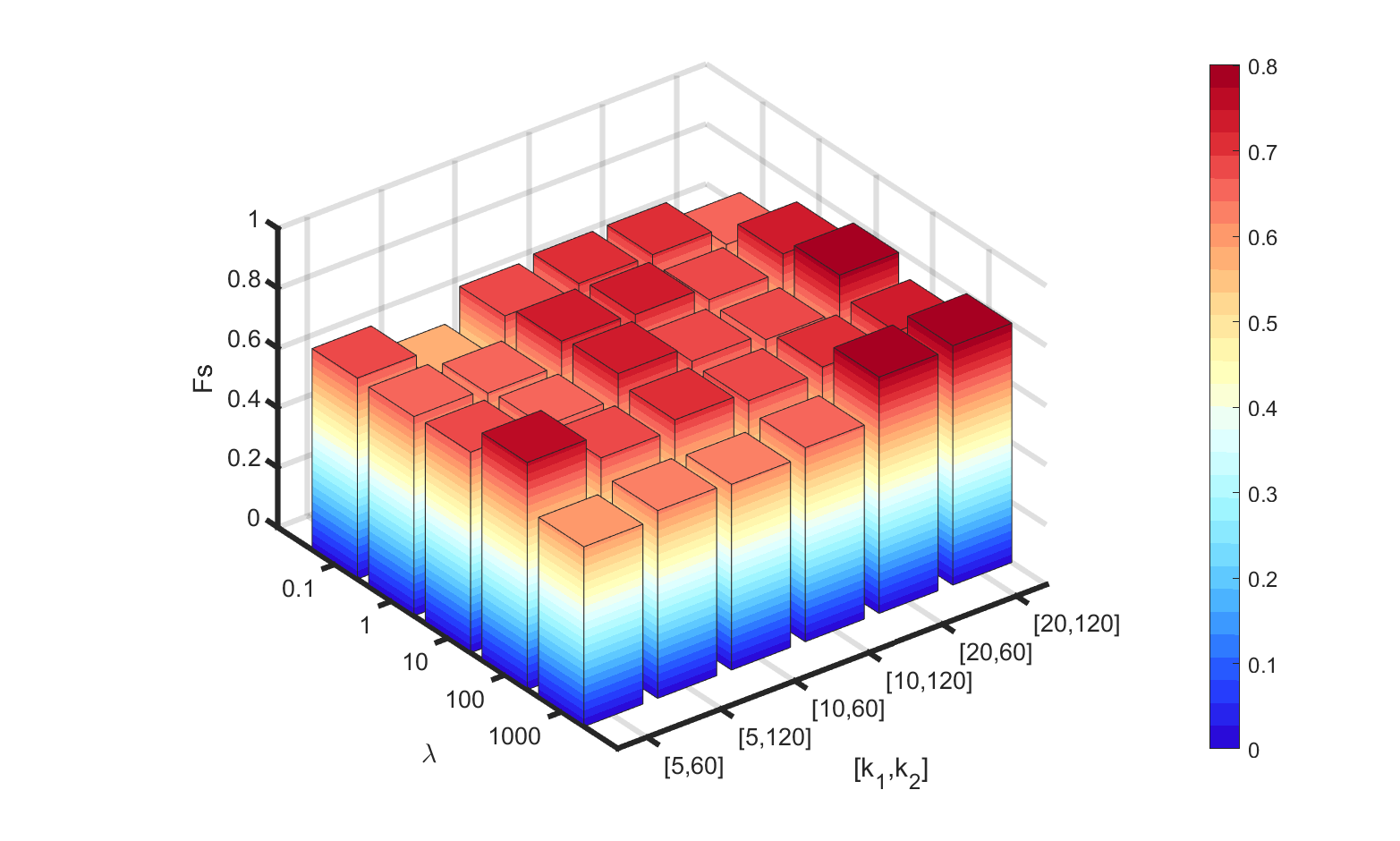}}
    \label{para3sources-c}}
    \subfigure[ARI]{
    \resizebox*{3.4cm}{!}{\includegraphics[scale=0.2]{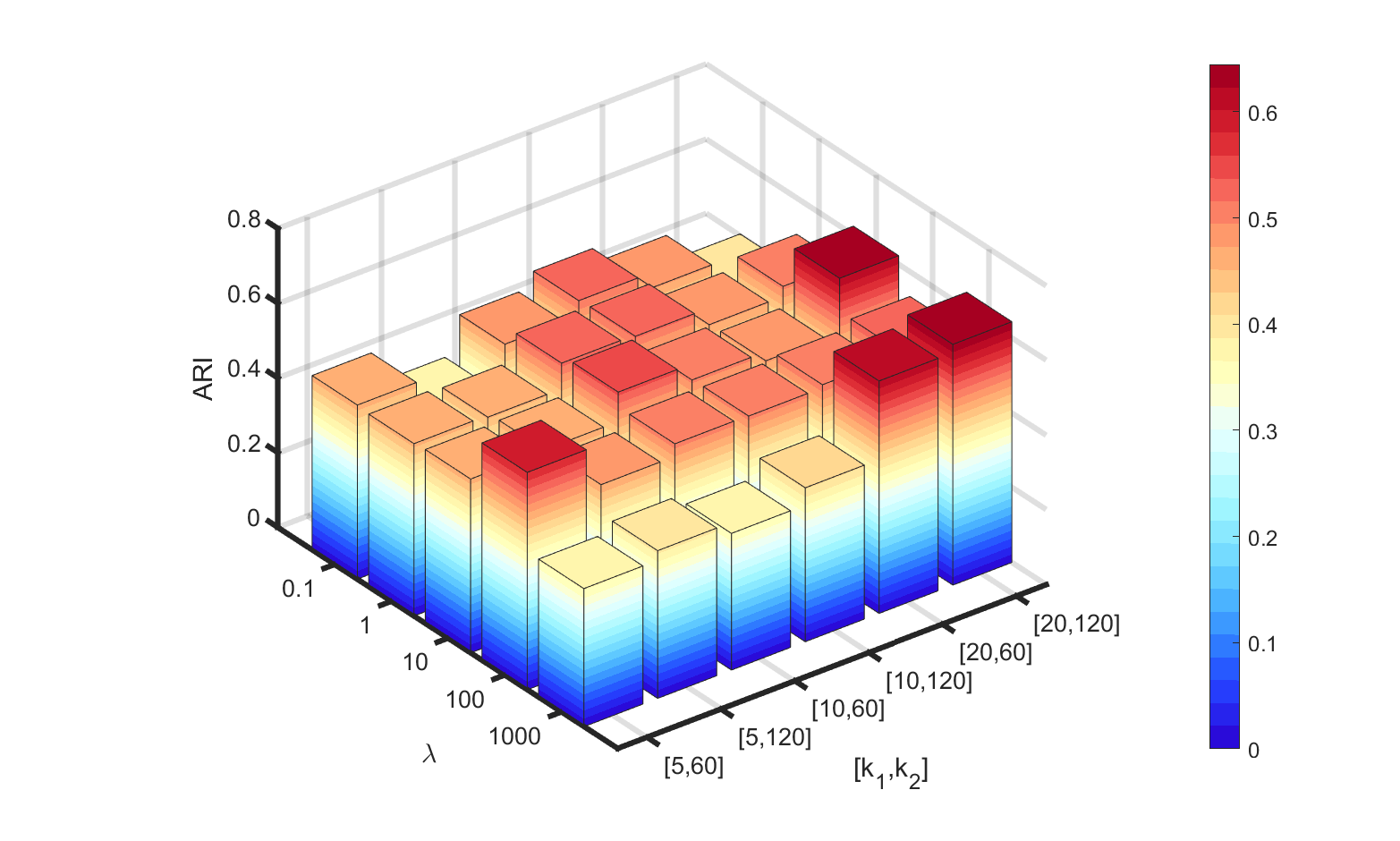}}
    \label{para3sources-d}}
    \caption{Parameter effects on 3sources.}
    \label{para3sources}
\end{figure}


\begin{figure}[H]
    \centering
    \subfigure[ACC]{
    \resizebox*{3.4cm}{!}{\includegraphics[scale=0.2]{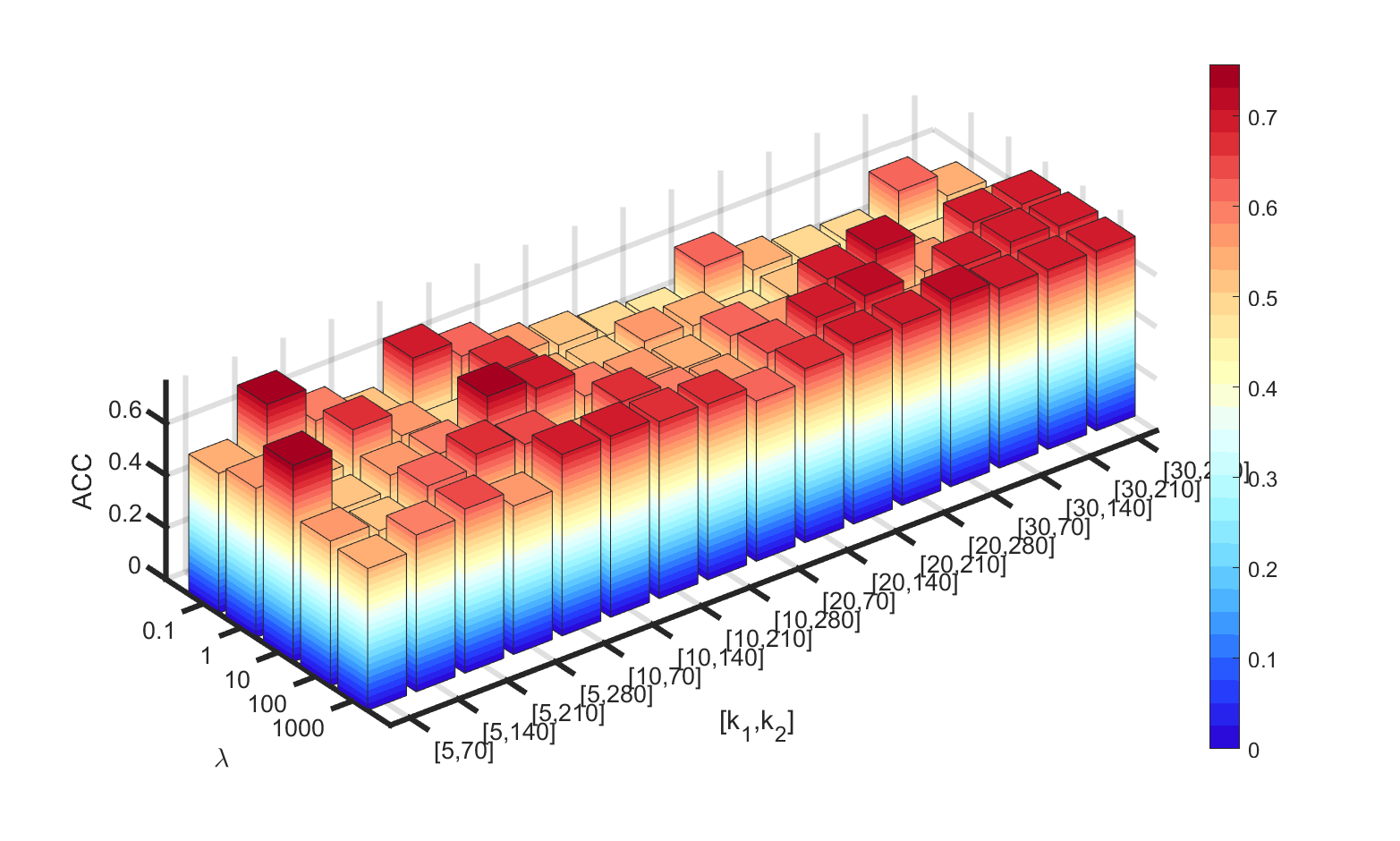}}
    \label{paracal1017-a}}
    \subfigure[NMI]{
    \resizebox*{3.4cm}{!}{\includegraphics[scale=0.2]{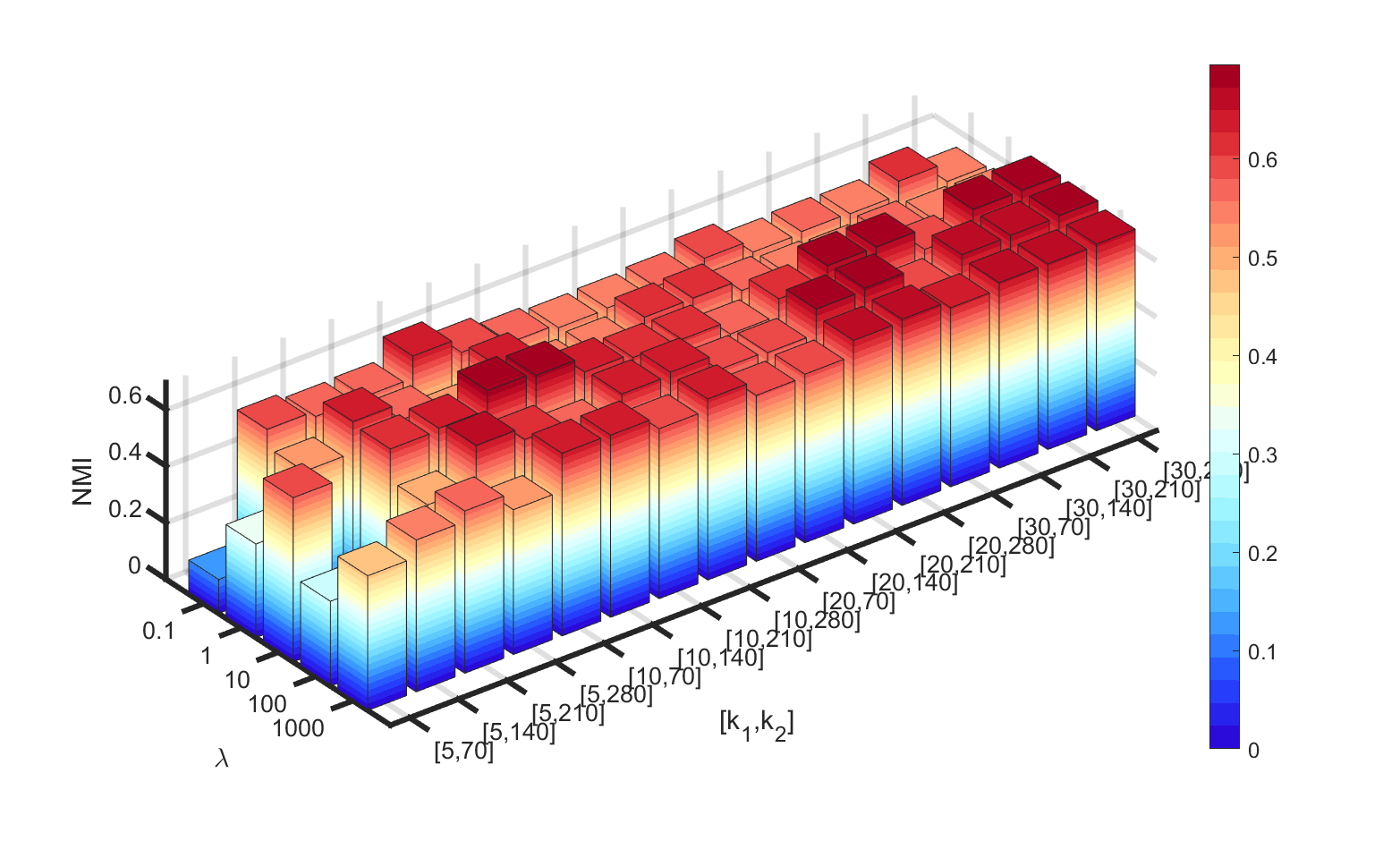}}
    \label{paracal1017-b}} \\
    \subfigure[Fs]{
    \resizebox*{3.4cm}{!}{\includegraphics[scale=0.2]{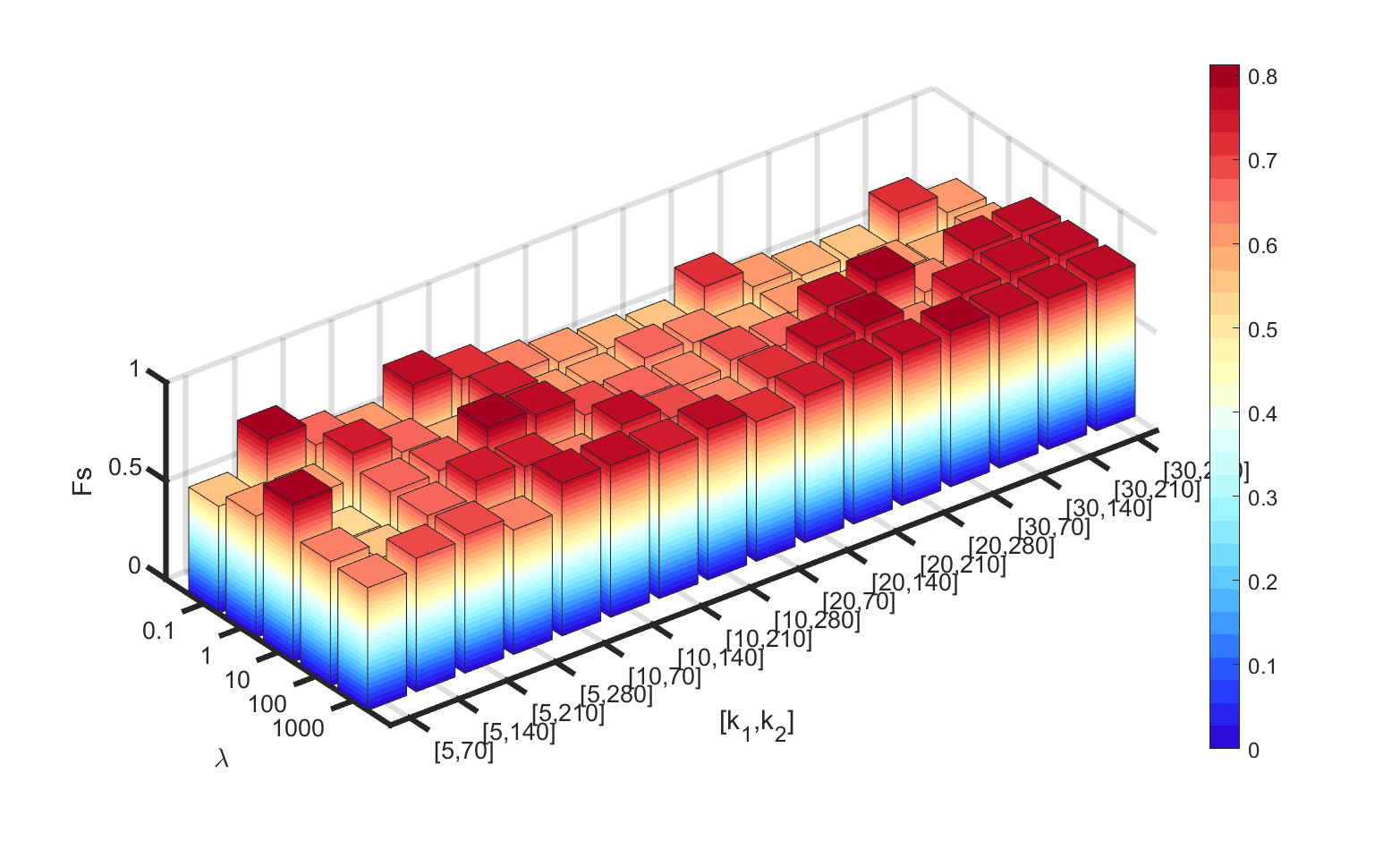}}
    \label{paracal1017-c}}
    \subfigure[ARI]{
    \resizebox*{3.4cm}{!}{\includegraphics[scale=0.2]{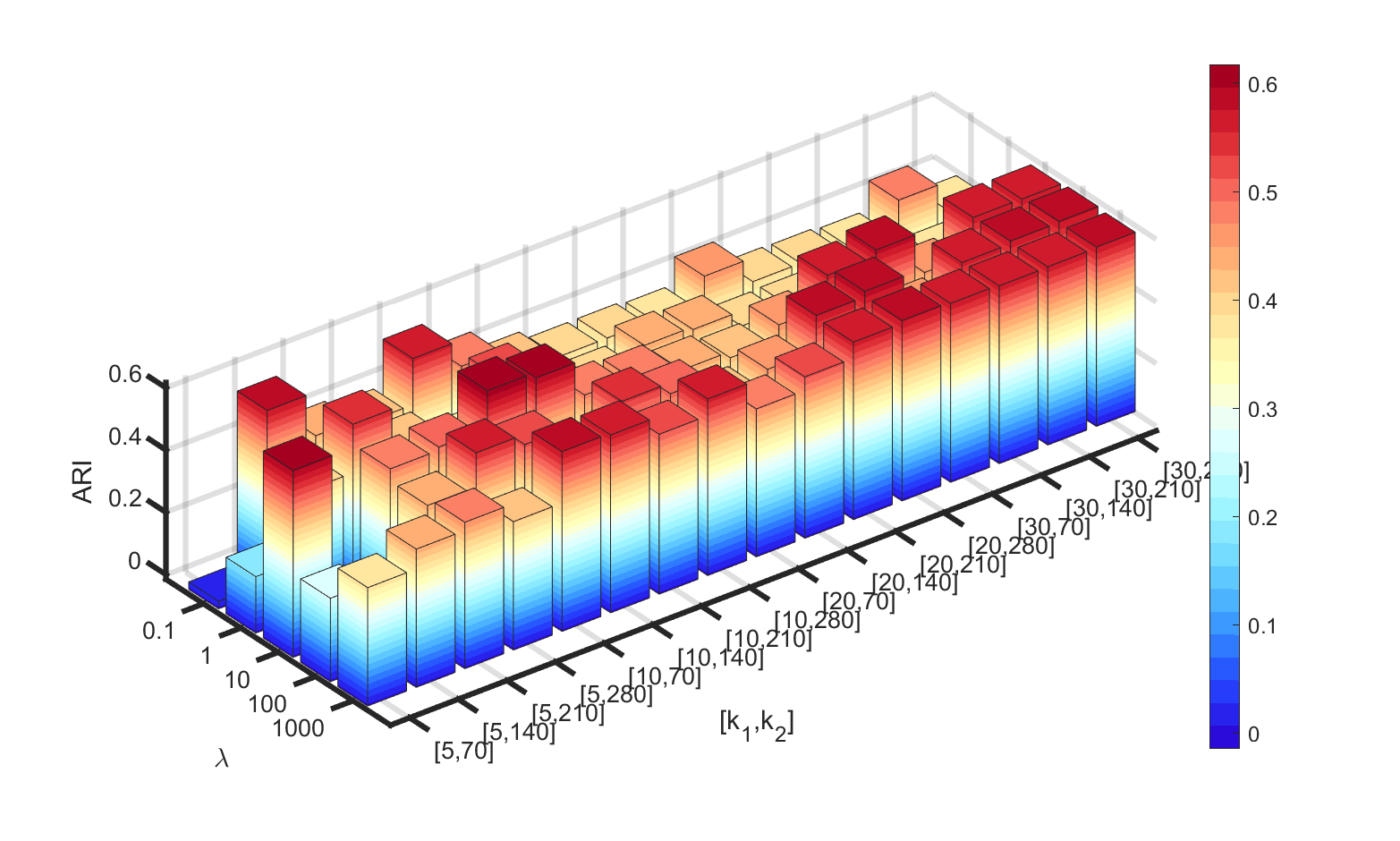}}
    \label{paracal1017-d}}
    \caption{Parameter effects on Caltech101-7.}
    \label{paracal1017}
\end{figure}

\vspace{-0.7cm}

\begin{figure}[H]
    \centering
    \subfigure[ACC]{
    \resizebox*{3.4cm}{!}{\includegraphics[scale=0.2]{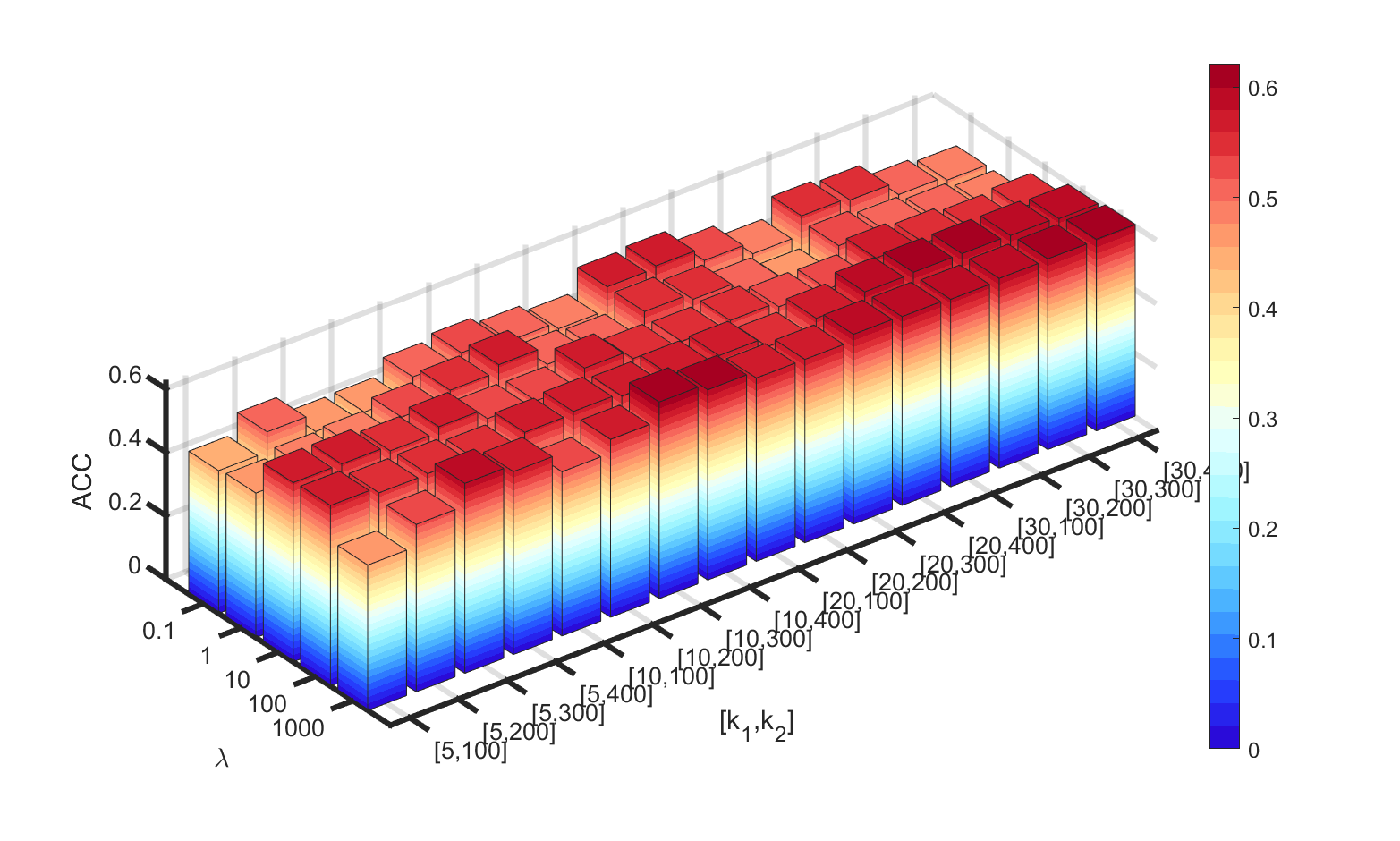}}
    \label{parawikipedia-a}}
    \subfigure[NMI]{
    \resizebox*{3.4cm}{!}{\includegraphics[scale=0.2]{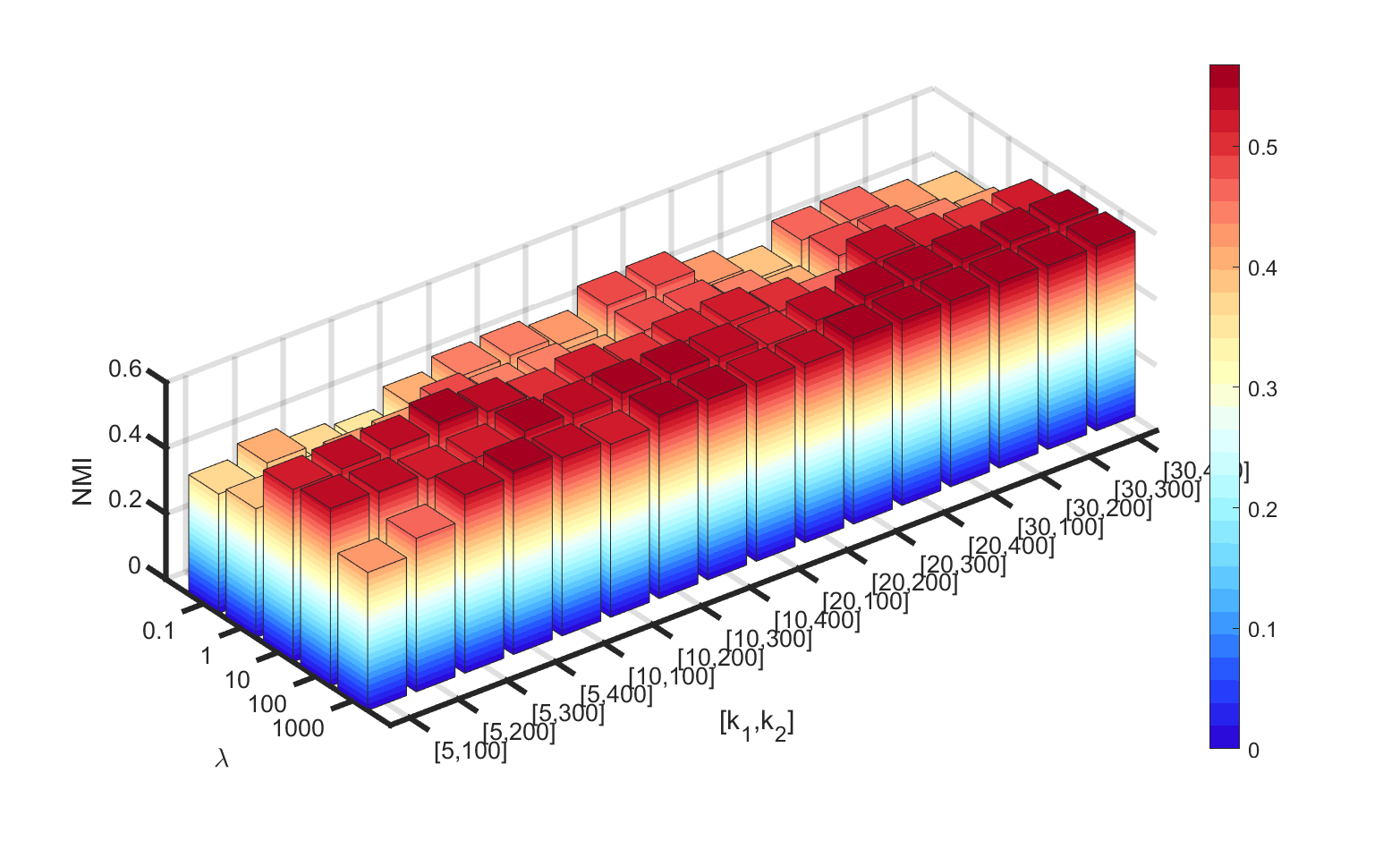}}
    \label{parawikipedia-b}} \\
    \subfigure[Fs]{
    \resizebox*{3.4cm}{!}{\includegraphics[scale=0.2]{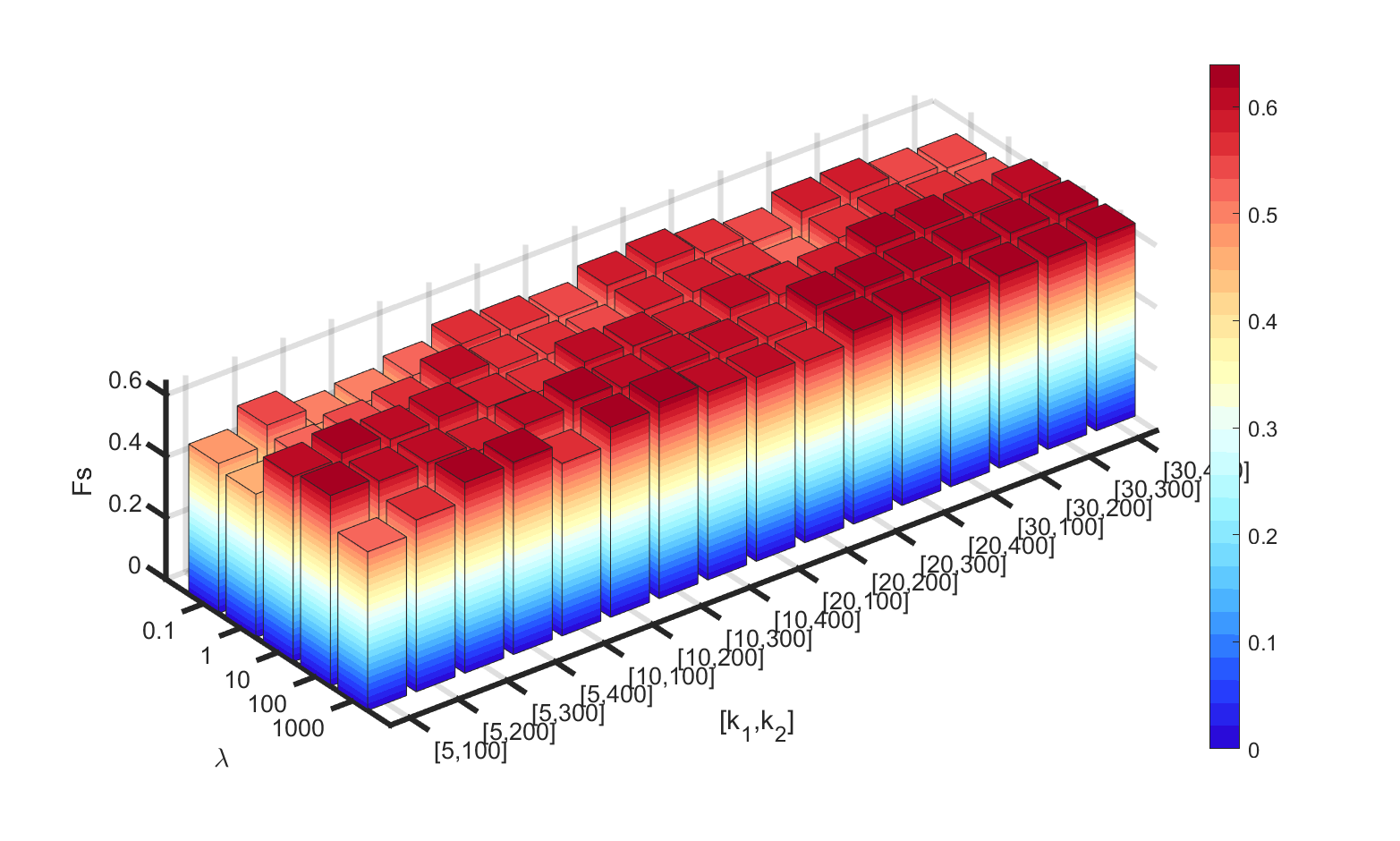}}
    \label{parawikipedia-c}}
    \subfigure[ARI]{
    \resizebox*{3.4cm}{!}{\includegraphics[scale=0.2]{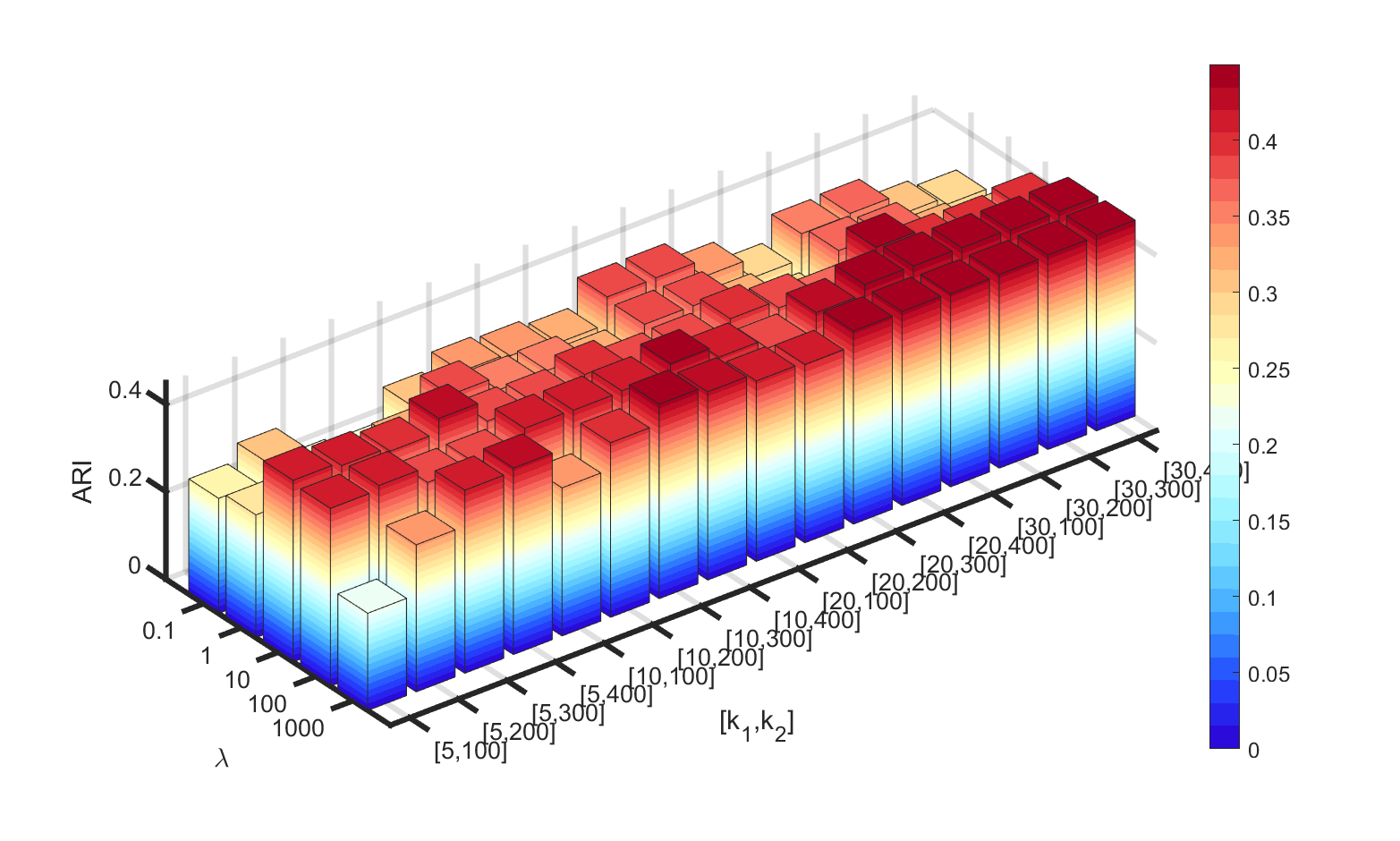}}
    \label{parawikipedia-d}}
    \caption{Parameter effects on wikipedia.}
    \label{parawikipedia}
\end{figure}

\vspace{-0.7cm}

\begin{figure}[H]
    \centering
    \subfigure[ACC]{
    \resizebox*{3.4cm}{!}{\includegraphics[scale=0.2]{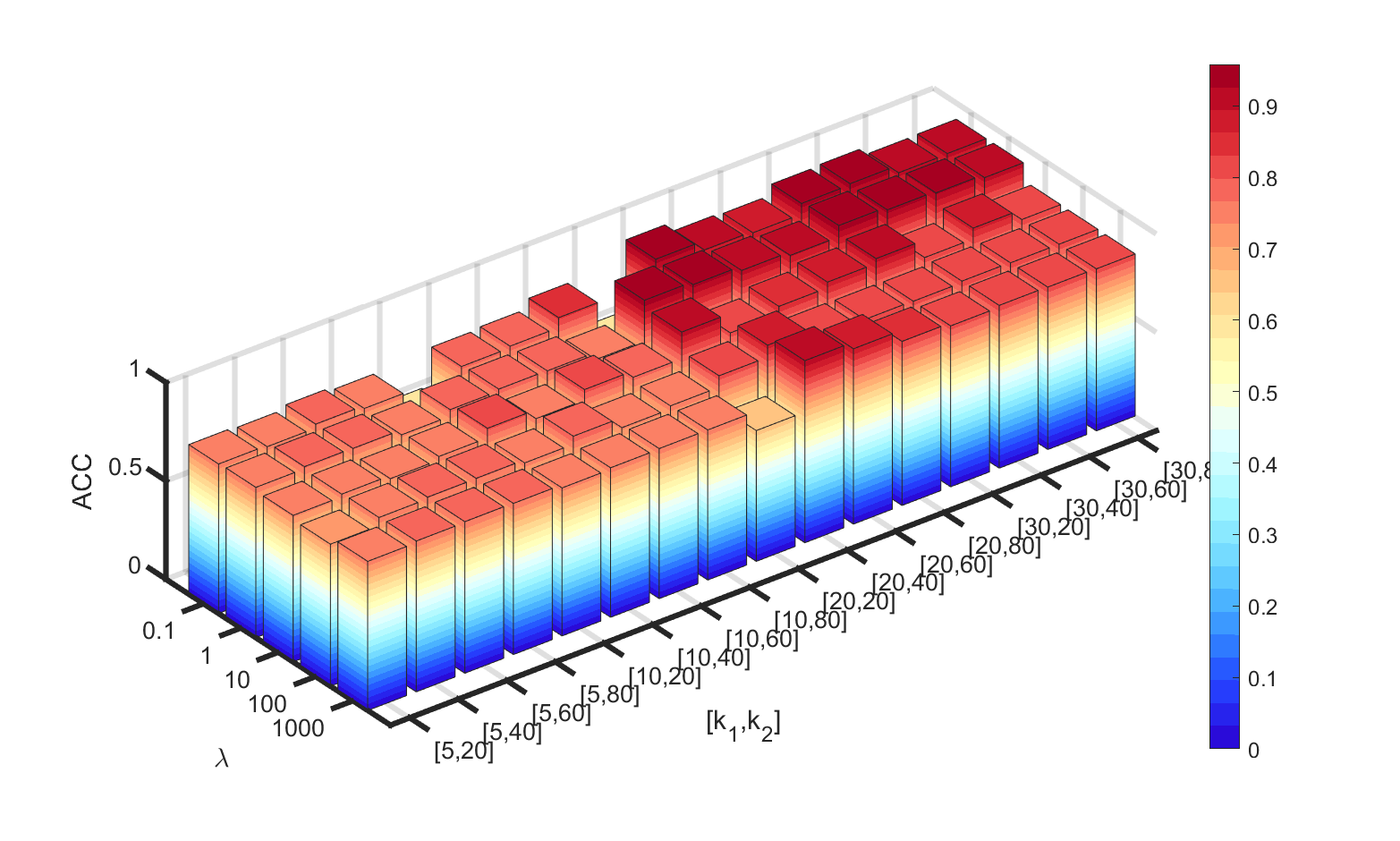}}
    \label{parawebkb-a}}
    \subfigure[NMI]{
    \resizebox*{3.4cm}{!}{\includegraphics[scale=0.2]{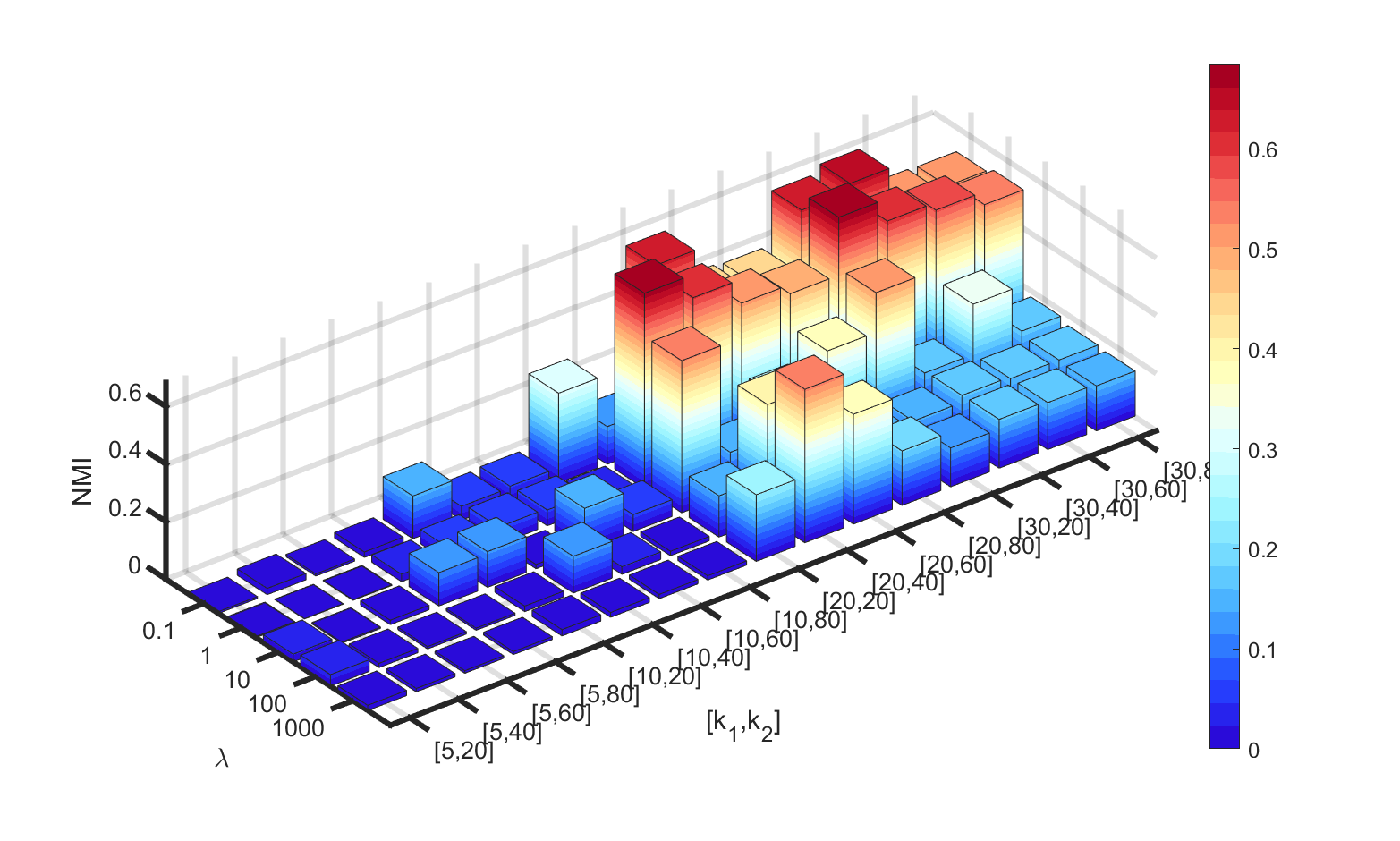}}
    \label{parawebkb-b}} \\
    \subfigure[Fs]{
    \resizebox*{3.4cm}{!}{\includegraphics[scale=0.2]{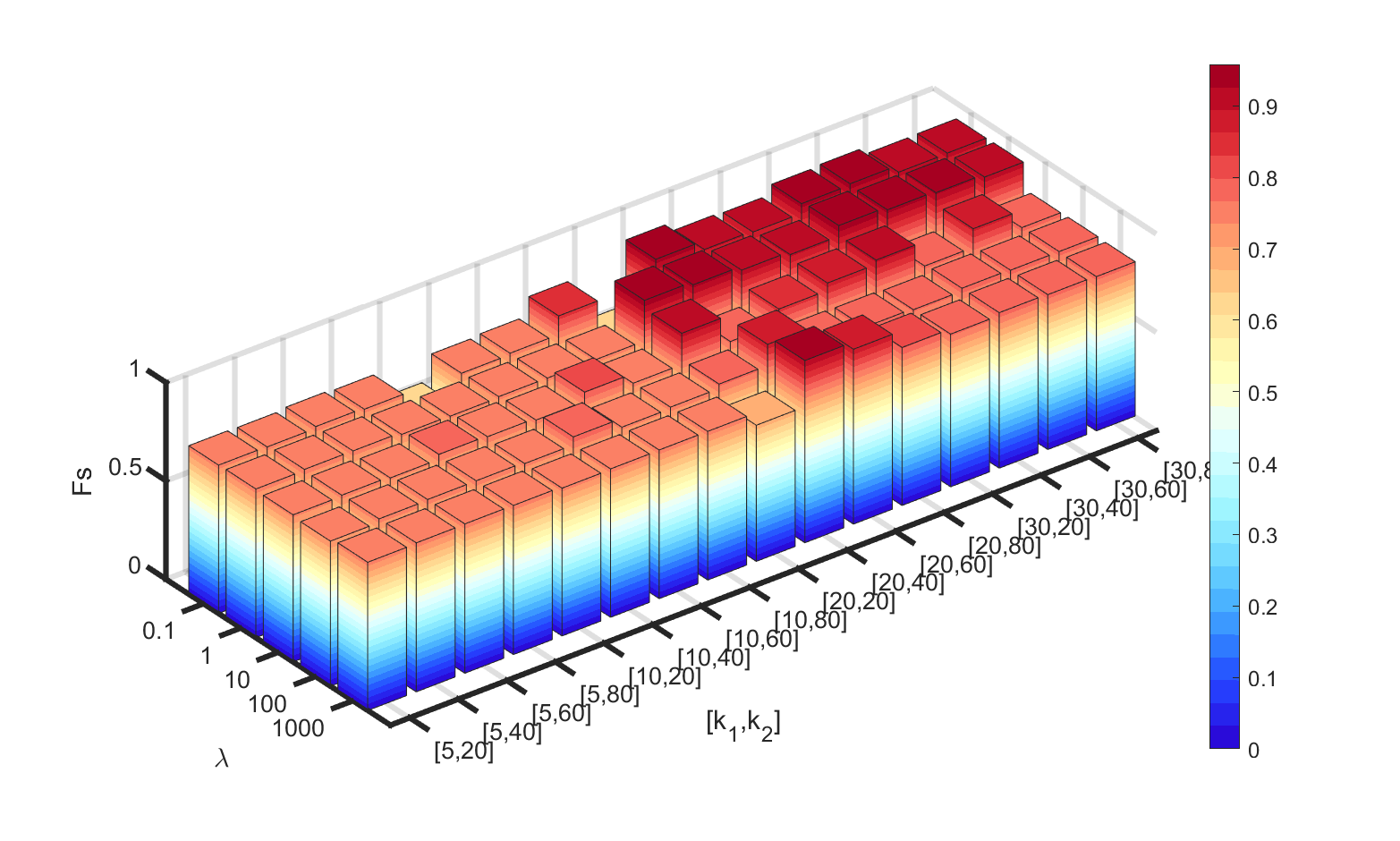}}
    \label{parawebkb-c}}
    \subfigure[ARI]{
    \resizebox*{3.4cm}{!}{\includegraphics[scale=0.2]{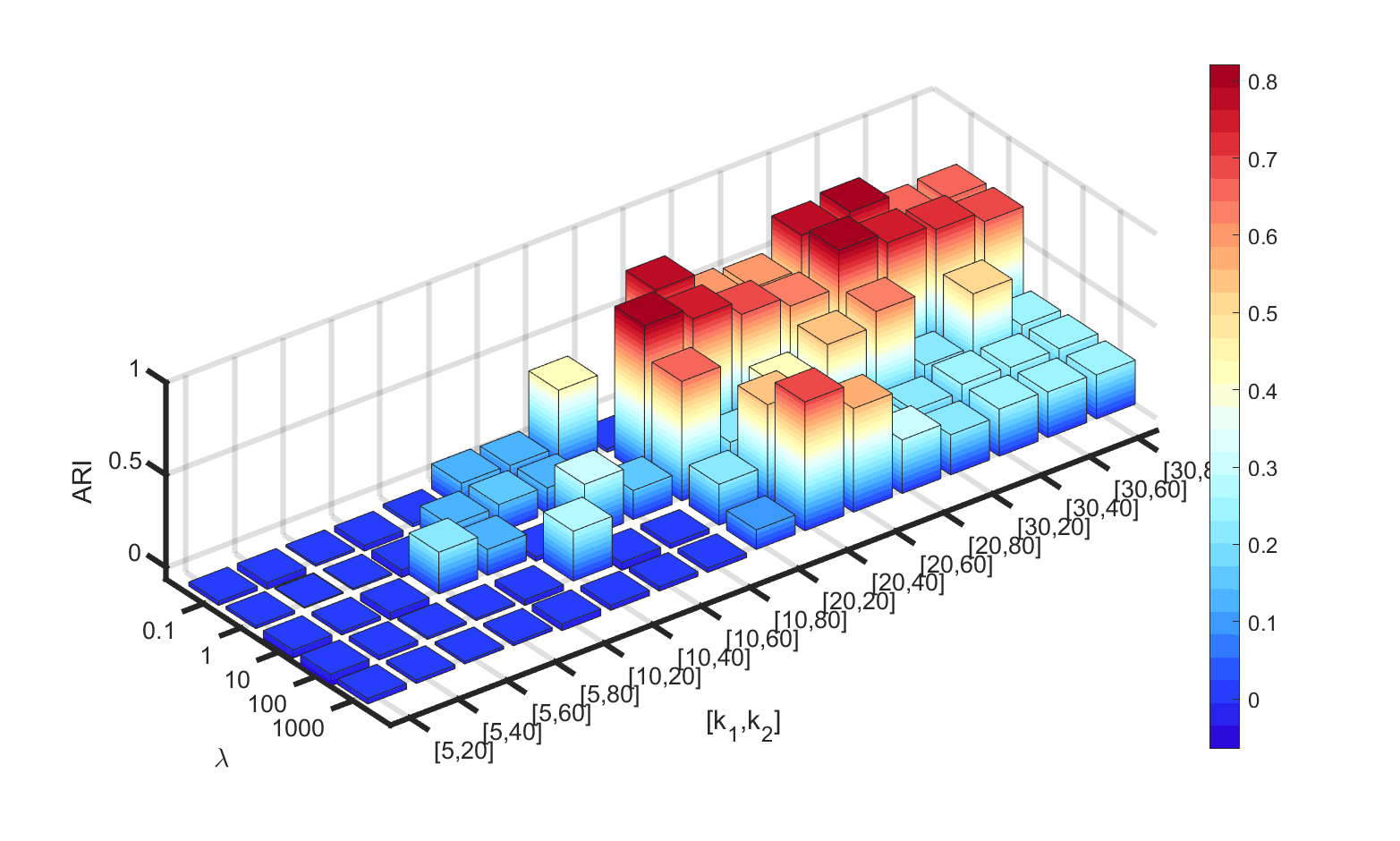}}
    \label{parawebkb-d}}
    \caption{Parameter effects on WebKB.}
    \label{parawebkb}
\end{figure}




\end{appendices}


\bibliography{sn-bibliography}

\end{document}